\documentclass{article} 
\usepackage{iclr2027_conference,times}

\usepackage{amsmath,amsfonts,bm}

\def\eqref#1{equation~\ref{#1}}

\def\1{\bm{1}}

\DeclareMathAlphabet{\mathsfit}{\encodingdefault}{\sfdefault}{m}{sl}
\SetMathAlphabet{\mathsfit}{bold}{\encodingdefault}{\sfdefault}{bx}{n}

\usepackage{hyperref}
\usepackage{url}

\usepackage{graphicx}
\usepackage{array}
\usepackage{tabularx}
\usepackage{longtable}
\usepackage{multirow}
\usepackage{booktabs}
\usepackage{xspace}
\usepackage{algorithm}
\usepackage{algpseudocode}
\usepackage{amsmath,amssymb}
\usepackage{wrapfig}
\usepackage{subcaption}
\usepackage{enumitem}
\usepackage{tikz}
\usepackage{xcolor}
\usepackage[percent]{overpic}
\usepackage{listings}
\usepackage{tcolorbox}
\tcbuselibrary{listings,breakable,skins}

\lstdefinestyle{promptstyle}{
    basicstyle=\ttfamily\footnotesize,
    columns=fullflexible,
    keepspaces=true,
    breaklines=true,
    breakatwhitespace=true,
    showstringspaces=false,
    aboveskip=0pt,
    belowskip=0pt,
    upquote=true,
    literate={---}{{\textemdash}}1,
}
\newtcblisting{promptbox}[1]{
    enhanced,
    breakable,
    listing only,
    listing engine=listings,
    listing options={style=promptstyle},
    colback=white,
    colframe=black!50,
    boxrule=0.4pt,
    arc=1.5mm,
    outer arc=1.5mm,
    left=2mm,
    right=2mm,
    top=1.5mm,
    bottom=1.5mm,
    title={#1},
    colbacktitle=black,
    coltitle=white,
    fonttitle=\bfseries\small,
    center title,
    titlerule=0pt,
    before skip=8pt,
    after skip=8pt,
}

\newcommand{\bcirc}[1]{%
  \tikz[baseline=(char.base)]{
    \node[shape=circle, fill=black, inner sep=1.2pt] (char)
      {\color{white}\scriptsize\bfseries #1};
  }%
}

\usepackage{array}
\usepackage{colortbl}

\newcolumntype{G}{>{\columncolor{gray!10}}c}

\newcolumntype{D}{c}

\newcommand{\framework}{ParetoSaddler\xspace}

\title{ParetoSaddler: Pareto-Guided Optimization of Agent Harnesses for Quality and Cost}

\author{
Xingzhi Qian\textsuperscript{1}\thanks{Work done during an internship at Microsoft.} \quad
Sungho Park\textsuperscript{2}\footnotemark[1] \quad
Rongyuan Tan\textsuperscript{3}\footnotemark[1] \quad
Jianbo Yu\textsuperscript{4}\footnotemark[1] \\
~\textbf{Jue Zhang}\textsuperscript{5}\thanks{Corresponding author.} \quad
\textbf{Lorenzo Cavallaro}\textsuperscript{6,1} \quad
\textbf{Menglin Xia}\textsuperscript{5} \quad
\textbf{Yanzhi Chen}\textsuperscript{5} \\
~\textbf{Xinnuo Xu}\textsuperscript{5} \quad
\textbf{Yu Kang}\textsuperscript{5} \quad
\textbf{Xinran Zheng}\textsuperscript{1} \quad
\textbf{Gagan Somashekar}\textsuperscript{5} \\
~\textbf{Xuchao Zhang}\textsuperscript{5} \quad
\textbf{Chetan Bansal}\textsuperscript{5} \quad
\textbf{Qingwei Lin}\textsuperscript{5} \quad
\textbf{Victor R\"{u}hle}\textsuperscript{5} \\[0.5em]
\textsuperscript{1}University College London \quad
\textsuperscript{2}Pohang University of Science and Technology \\
\textsuperscript{3}Southern University of Science and Technology \quad
\textsuperscript{4}Southeast University \\
\textsuperscript{5}Microsoft \quad
\textsuperscript{6}Mohamed bin Zayed University of Artificial Intelligence \\
\texttt{juezhang@microsoft.com}
}

\newif\ifshowcomments
\showcommentstrue 
\ifshowcomments
    \newcommand{\jz}[1]{{\color{blue}[JZ: #1]}}
    \newcommand{\xq}[1]{{\color{purple}[XQ: #1]}}
    
\else
    \newcommand{\jz}[1]{}
    \newcommand{\xq}[1]{}
\fi

\iclrpreprintcopy
\begin{document}

\maketitle

\begin{abstract}


Harness optimization can substantially improve LLM agents without modifying model weights, but existing methods primarily optimize response quality. Extending them to joint quality--cost optimization introduces a challenge: applying the deployment preference during intermediate updates can prematurely reject candidates that are valuable for later improvements. We introduce ParetoSaddler, which addresses this mismatch by separating intermediate candidate promotion from final deployment selection. At each iteration, an optimizer updates a parent harness to produce a candidate, and then compares the two over repeated executions and uses Bayesian bootstrap Pareto support to promote candidates showing credible progress in either quality or cost. Promoted candidates receive broader development evaluation and remain available as parents for later optimization steps. We evaluate ParetoSaddler on ClawMark and GAIA2 using domain-disjoint splits. On ClawMark, ParetoSaddler improves quality from 70.86\% to 81.35\% while reducing cost from \$1.17 to \$1.01 per task; on GAIA2, it improves quality
from 56.22\% to 62.89\% while reducing cost from \$0.78 to \$0.45 per task. The Pareto-selected harnesses also achieve higher quality at lower cost than quality-only harness optimization. Analysis shows that Pareto-aware optimization retains intermediate candidates that would be discarded by deployment preference but can support later improvements, while Bayesian bootstrap screening filters out candidates whose gains are weakly supported across runs. Project website and code will be available at \url{https://aka.ms/ParetoSaddler-website}.

\end{abstract}

\begin{figure*}[h]
   \centering

    \includegraphics[width=\linewidth]{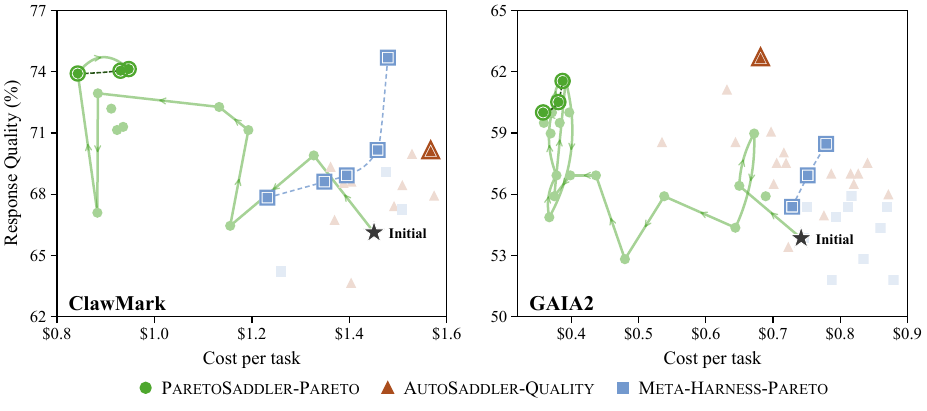}
    \caption{\textbf{Quality--cost optimization on the development sets of ClawMark and GAIA2.}
    Each marker represents a harness evaluated on the development set;
    better solutions lie toward the upper left. Stars denote the initial harness, and enlarged outlined markers highlight the final candidates
    retained by each method.
    Solid green lines trace harness evolution in ParetoSaddler; green markers not connected by solid lines denote candidates
    not selected for further refinement.
    Dashed lines connect candidates on the Pareto fronts of ParetoSaddler and Meta-Harness.
    ParetoSaddler discovers high-quality harnesses in a lower-cost
    region not reached by the evaluated baseline candidates.
    }
    \label{fig:dev_pareto}
\end{figure*}

\section{Introduction}

Large language models (LLMs) have enabled agents that reason, use tools, and interact with executable environments to solve increasingly complex tasks. An agent's capabilities arise from both its underlying model and its \emph{harness}: the prompts, tools, middleware, and control logic that mediate its interaction with the environment. Even with model weights fixed, modifying these components can substantially improve agent performance~\citep{langchain-harness-practice}. This has motivated growing interest in automatic harness optimization (AHO)~\citep{lee2026metaharness,park2026autosaddler}, in which an external optimizer diagnoses execution traces and updates the surrounding harness to improve the task agent.

Most existing AHO methods optimize primarily for task quality, whereas real-world deployment must also consider execution cost and other requirements such as safety and privacy. Here, we focus on jointly optimizing task response quality and LLM token cost. Deployment preferences may specify a fixed quality--cost trade-off, impose an explicit constraint, or defer selection among Pareto alternatives. This multi-objective setting raises a key question beyond final deployment selection: \emph{which intermediate harness updates should remain available for continued optimization?}

This distinction is particularly important for AHO methods with
\emph{staged evaluation}, where locally generated updates are screened before
receiving broader development evaluation~\citep{park2026autosaddler}. The
screening decision does more than reduce evaluation cost: it also determines
which intermediate harnesses remain available for subsequent refinement.
Applying the final deployment preference directly at this stage can therefore
prematurely reject updates whose immediate quality--cost trade-off is
unfavorable but whose changes enable stronger solutions later, consistent with
prior optimization research showing that search guidance need not coincide
with the ultimate objective~\citep{knowles2001multiobjectivization,lehman2011novelty}.
For long-horizon agents, this mismatch can arise naturally because execution
cost is often coupled with task progress: a harness may appear inexpensive
simply because some executions terminate prematurely, whereas a patch that
fixes such failures may improve quality while increasing cost through
additional reasoning or environment interaction. Rejecting the update solely
because it temporarily violates the deployment cost preference can prevent
later patches from building on its capability gains and reducing the newly
exposed inefficiencies.

Local promotion is further complicated by stochastic execution. Repeated runs
of the same harness can vary substantially in both quality and cost, so an
apparent improvement may reflect execution randomness rather than the update
itself. Simply comparing mean outcomes collapses this
variation into a point estimate and can therefore overstate weakly supported gains. Staged multi-objective AHO thus faces two questions: \emph{what intermediate movements are worth preserving for further
optimization, and how can we assess whether their observed gains are
statistically supported?}

To address these issues, we introduce~\textbf{\framework}, a framework for staged quality--cost harness
optimization that separates
\emph{exploration-time promotion} from \emph{deployment-time selection}.
At each iteration, an underlying AHO method constructs a candidate harness from
execution evidence. \framework repeatedly evaluates the candidate and its
parent and uses Bayesian bootstrap support~\citep{rubin1981bayesian} to promote
updates showing credible progress in either quality or cost, without requiring
them to already satisfy the final deployment preference. 
Only after the optimization
budget is exhausted is that preference applied to select the
final harness. In this way, \framework separates what is
worth \emph{exploring} during optimization from what is ultimately worth
\emph{deploying}.

We evaluate \framework on
ClawMark~\citep{meng2026clawmark} and GAIA2~\citep{froger2026gaia2} using domain-disjoint training,
development, and test splits under both Pareto-set optimization and
quality-constrained cost minimization. 
On ClawMark, ParetoSaddler improves quality on the held-out
test set from 70.86\% to 81.35\% while reducing cost from
\$1.17 to \$1.01 per task; on GAIA2, quality improves from
56.22\% to 62.89\% while cost decreases from \$0.78 to \$0.45. Notably, its quality-oriented Pareto solution
also outperforms quality-only harness optimization by 7.49 points
on ClawMark and 5.56 points on GAIA2 while remaining cheaper, showing that jointly
optimizing quality and cost need not sacrifice quality and can even lead the
search to stronger quality solutions.

Our main contributions are summarized as follows:
\begin{itemize}[leftmargin=1em,itemsep=0.25em,topsep=0.25em]
    \item When performing multi-objective agent harness optimization with staged evaluation, we identify a mismatch between \emph{intermediate promotion} and
    \emph{final deployment selection}, where enforcing deployment preferences
    too early can prune updates useful for later optimization.

    \item We introduce \framework, which combines Pareto-aware promotion, repeated
    execution, and Bayesian bootstrap screening for effective and reliable quality--cost
    harness optimization.

    \item We demonstrate consistent quality--cost improvements on ClawMark and
    GAIA2, supported by detailed analyses and ablations of
promotion, repeated execution, and Bayesian screening.
\end{itemize}

\section{Related Work}

\paragraph{Automatic Harness Optimization.}
Automatic optimization of LLM systems has expanded from prompts
\citep{pryzant2023apo, yuksekgonul2024textgrad, guo2024evoprompt, yang2024opro, agrawal2026gepa} and reusable skills~\citep{yang2026skillopt, ni2026trace2skill, alzubi2026evoskill} to complete agent harnesses.
Existing AHO methods broadly follow two optimization protocols. \emph{Search-based} approaches~\citep{lou2026autoharness,lee2026metaharness,pan2026rho} generate and compare harness candidates using execution feedback, iteratively refining promising candidates.
\emph{Learning-based} approaches~\citep{park2026autosaddler,jiang2026harnessevolve} derive incremental updates from mini-batches and use staged evaluation
to decide which updates to retain.
These protocols are orthogonal to whether adaptation occurs online during deployment or offline on a fixed development corpus.
ParetoSaddler follows the staged offline-learning setting and focuses on joint quality--cost optimization.

\paragraph{Multi-Objective Optimization for LLM Systems.}
Classical multi-objective optimization retains non-dominated solutions
representing different trade-offs~\citep{knowles2000paes, zitzler2001spea2}, and recent work applies this principle to prompts~\citep{yang2023instoptima, sinha2024survival, bussing2026mocapo}
and agent skills~\citep{gong2026skillmoo}.
Full-harness optimization is more costly because evaluating each
candidate requires long agent trajectories, motivating staged
evaluation that reserves broader evaluation for promising candidates
~\citep{schmucker2021multiobjectiveasynchronoussuccessivehalving}. 
Prior work shows that search guidance need not coincide with the final selection objective
~\citep{knowles2001multiobjectivization,lehman2011novelty}. Building on this distinction, ParetoSaddler uses Pareto-aware promotion during optimization while reserving the deployment preference for final harness selection.

\paragraph{Optimization under Stochastic Evaluation.}
Candidate selection under noisy evaluations is a longstanding challenge in
optimization~\citep{branke2003selection, marceau2014racing}. 
Recent work has applied repeated evaluation and bootstrap resampling to multi-objective ranking and LLM-agent configuration selection
~\citep{rook2024multiobjective, vattikonda2025train}. ParetoSaddler builds on this by incorporating Bayesian bootstrap uncertainty estimation~\citep{rubin1981bayesian} into staged AHO, using statistical support to determine candidate promotion.

\section{Preliminaries and Problem Formulation}
\label{sec:formulation}

\paragraph{Agent Harness and Stochastic Execution.}
An agent harness comprises the prompts, tools, middleware, and control logic surrounding the underlying LLM~\citep{zhou2026externalization}. 
Let $h\in\mathcal H$ denote a candidate harness and $m$ the underlying LLM, which remains fixed throughout optimization. Executing the agent $(h,m)$ on task $x$ induces a stochastic trajectory
$
\tau \sim p(\tau\mid h,m,x),
$
with randomness arising from model generation and interactions with tools and the environment.

\paragraph{Quality and Cost Objectives.}
We jointly optimize task performance and LLM token cost. Hereafter,
we use \emph{quality} as a generic term for benchmark-specific task
performance, which may be measured by task success or rubric-based scoring, and \emph{cost} to denote the monetary cost of LLM
token usage under a fixed price card. Each trajectory therefore yields a
normalized quality score \(q(\tau)\in[0,1]\), with higher values indicating
better task performance, and incurs a cost \(c(\tau)\geq 0\).

For the \(j\)-th call, let
\(T_{j,\mathrm{in}}\), \(T_{j,\mathrm{cache}}\), and
\(T_{j,\mathrm{out}}\) denote the numbers of fresh-input, cached-input,
and output tokens, respectively, with corresponding per-token prices
\(\rho_{\mathrm{in}}\), \(\rho_{\mathrm{cache}}\), and
\(\rho_{\mathrm{out}}\). For a trajectory \(\tau\) consisting of
\(M(\tau)\) model calls, its total cost is then defined as 
$
c(\tau)=\sum_{j=1}^{M(\tau)}
\left(
\rho_{\mathrm{in}}T_{j,\mathrm{in}}
+\rho_{\mathrm{cache}}T_{j,\mathrm{cache}}
+\rho_{\mathrm{out}}T_{j,\mathrm{out}}
\right)$.
For a target task distribution $\mathcal P$, the population-level quality and cost of harness $h$ are
\begin{equation}
Q(h)
=
\mathbb E_{x\sim\mathcal P,
\tau\sim p(\cdot\mid h,m,x)}[q(\tau)],
\qquad
C(h)
=
\mathbb E_{x\sim\mathcal P,
\tau\sim p(\cdot\mid h,m,x)}[c(\tau)].
\label{eq:population_objectives}
\end{equation}
We seek harnesses with high $Q(h)$ and low $C(h)$. In practice, we estimate these expectations using a finite task dataset $\mathcal D$. Given $R$ independent executions per task, let $\tau_{i,r}^{h}$ denote the $r$-th trajectory of harness $h$ on task $x_i$. We average executions within each task and then weight tasks equally:
\begin{equation}
\widehat Q_{\mathcal D}(h)
=
\frac{1}{|\mathcal D|}
\sum_{i=1}^{|\mathcal D|}
\frac{1}{R}\sum_{r=1}^{R}q(\tau_{i,r}^{h}),
\qquad
\widehat C_{\mathcal D}(h)
=
\frac{1}{|\mathcal D|}
\sum_{i=1}^{|\mathcal D|}
\frac{1}{R}\sum_{r=1}^{R}c(\tau_{i,r}^{h}).
\label{eq:empirical_objectives}
\end{equation}
Optimization decisions use these empirical estimates on the designated data split.

\paragraph{Deployment Preferences and Optimization Formulations.}
A deployment preference $\pi$ specifies which quality--cost trade-off is desirable. Depending on whether these preferences are expressed through fixed weights, explicit constraints, or deferred selection among alternatives, we have:

\begin{itemize}[leftmargin=1em,itemsep=0.25em,topsep=0.25em]

\item \textbf{Weighted-Sum Scalarization}
combines quality and cost into a single objective:
\begin{equation}
    h_{\mathrm{weighted}}
    \in
    \arg\max_{h\in\mathcal H}
    \left[
        Q(h)-\lambda C(h)
    \right],
    \label{eq:weighted_objective}
\end{equation}
where $\lambda>0$ is a predefined weight balancing quality and cost.

\item \textbf{Constrained Optimization}
optimizes one objective subject to a bound on the other.
For example, we formulate \emph{quality-constrained cost minimization} as follows:
\begin{equation}
    h_{\mathrm{constrained}}
    \in
    \arg\min_{h\in\mathcal H} C(h)
    \quad
    \text{subject to}
    \quad
    Q(h)\geq Q_{\mathrm{min}},
    \label{eq:constrained_objective}
\end{equation}
where $Q_{\mathrm{min}}$ is the quality floor. Alternatively, one can maximize quality under a cost budget.

\item \textbf{Pareto-Set Optimization}
seeks a set of alternatives spanning quality--cost trade-offs when preferences are unspecified.
Harness $h_a$ dominates $h_b$, denoted by $h_a\succ h_b$, if
$
    Q(h_a)\geq Q(h_b)
    \mathrm{~and~} 
    C(h_a)\leq C(h_b),
    \label{eq:dominance}
$
with at least one strict inequality.
The Pareto-optimal set is
\begin{equation}
    \mathcal H^\star
    =
    \left\{
        h\in\mathcal H:
        \nexists h'\in\mathcal H
        \text{ such that } h'\succ h
    \right\}.
    \label{eq:pareto_set}
\end{equation}
In practice, the goal is to approximate this frontier with a diverse set of non-dominated harnesses.

\end{itemize}

\paragraph{Iterative Harness Optimization with Staged Evaluation.}

We consider iterative harness optimization in which new candidates build
on previously explored harnesses. We use the datasets \(\mathcal D_{\mathrm{train}}\),
\(\mathcal D_{\mathrm{dev}}\), and
\(\mathcal D_{\mathrm{test}}\) for deriving updates,
guiding harness selection, and final assessment, respectively.
Because thoroughly evaluating every generated candidate can be expensive,
recent prompt and harness optimization and agent self-improvement methods~\citep{
agrawal2026gepa,park2026autosaddler,zhang2026hyperagents}
often adopt \textit{staged evaluation}: candidates first undergo a
relatively inexpensive screening procedure, and only selected candidates
receive broader evaluation. We refer to the decision to
advance a candidate beyond this preliminary stage as \emph{promotion}.

\section{Proposed Method: \framework}
\label{sec:method}

\begin{figure}[t]
    \centering
    \includegraphics[width=\linewidth]{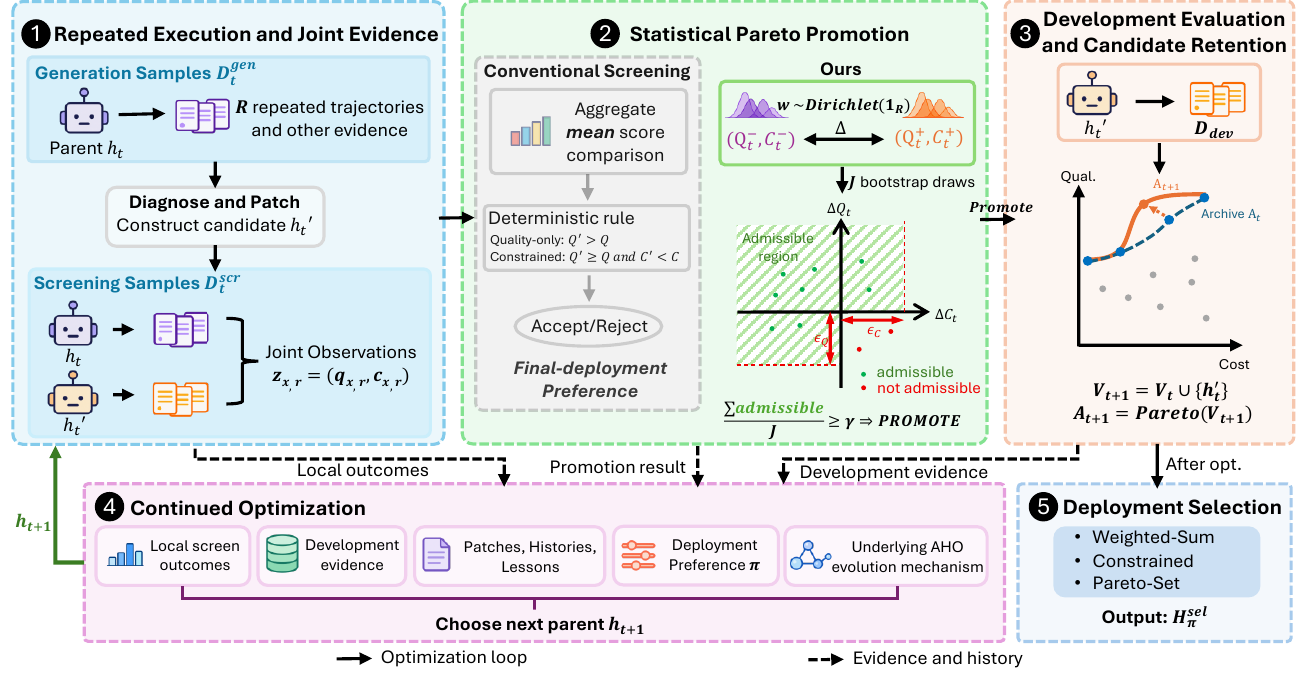}
    \caption{
    \textbf{Overview of \framework.}
    Repeated execution and statistical Pareto
    promotion identify candidates for development evaluation. The updated
    candidate pool and Pareto archive support continued optimization,
    with final harness selection governed by the deployment preference.
    }
    \label{fig:paretosaddler-overview}
\end{figure}

\paragraph{Overview.}
\framework brings Pareto reasoning into the promotion decisions of staged
harness optimization. As illustrated in
Figure~\ref{fig:paretosaddler-overview}, at iteration \(t\), the optimizer
selects a parent harness \(h_t\) and uses generation samples
\(\mathcal D_t^{\mathrm{gen}} \subseteq \mathcal D_{\mathrm{train}}\) to construct a candidate \(h'_t\)
(\bcirc{1}). These samples provide the execution traces and other
method-specific evidence used to derive the update. The parent and candidate are then assessed using screening samples
\(\mathcal D_t^{\mathrm{scr}}  \subseteq \mathcal D_{\mathrm{train}} \), which may differ from or overlap with the
generation samples. 
During screening, the parent and candidate are repeatedly executed under a shared evaluation protocol, yielding joint quality--cost observations.

A statistical Pareto promotion rule (\bcirc{2}) uses these observations to advance a candidate when there is sufficient evidence of improvement in either quality or cost, without requiring the update to satisfy a particular deployment preference. Promoted candidates then receive broader development evaluation (\bcirc{3}), which updates the development-evaluated alternatives and their empirical Pareto archive. The resulting alternatives, together with the deployment preference \(\pi\), guide continued optimization (\bcirc{4}), forming an iterative
loop. After the optimization budget is exhausted, the deployment preference \(\pi\) is applied to the retained candidates to determine the returned harness or harness set (\bcirc{5}). Algorithms~\ref{alg:paretosaddler-general}
and~\ref{alg:paretosaddler-autosaddler} provide the general procedure and its AutoSaddler instantiation, respectively.

\paragraph{Repeated Execution and Evaluation Evidence.}

Execution randomness affects both quality and cost. As shown in
Appendix~\ref{app:quality_cost_variations}, across 20 ClawMark tasks and six
repeated runs per task, the mean within-task coefficient of variation reaches
\textbf{18.8\%} for quality and \textbf{25.6\%} for cost. We therefore use
\(R\) repeated executions per task for both screening and development
evaluation.

Concretely, at iteration \(t\), the underlying AHO method executes the parent harness on the generation context \(\mathcal D_t^{\mathrm{gen}}\) and collects trajectories together with paired quality--cost observations. For an individual execution \(\tau\), we write \(\mathbf z=(q(\tau),c(\tau))\), while \(\mathcal Z\) denotes the collection of such observations across executions and tasks. This evidence, together with any method-specific information, is used to construct a candidate \(h'_t\). In our AutoSaddler instantiation, the collected traces are analyzed to diagnose task failures and inefficient behaviors and to propose targeted harness updates.

The candidate is then screened against its parent on \(\mathcal D_t^{\mathrm{scr}}\), using the same protocol and \(R\) executions per task for both harnesses. Let \(h_t^{-}=h_t\) and \(h_t^{+}=h'_t\), with screening evidence \((\mathcal T_t^{-},\mathcal Z_t^{-})\) and \((\mathcal T_t^{+},\mathcal Z_t^{+})\), respectively. When generation and screening use compatible tasks and protocols, parent executions can be reused. For example, AutoSaddler generates and screens updates on the same training mini-batch, so \(\mathcal D_t^{\mathrm{gen}}=\mathcal D_t^{\mathrm{scr}}\) and no additional parent executions are required.

\paragraph{Statistical Pareto Promotion.}

Repeated execution yields multiple quality--cost observations for each task and harness candidate, raising the question of how to consolidate them into a reliable promotion decision. Rather than relying on empirical means, we quantify uncertainty in the observed differences using the Bayesian bootstrap~\citep{rubin1981bayesian}. By assigning continuous Dirichlet weights to the repeated observations, the Bayesian bootstrap induces a nonparametric distribution over plausible quality--cost outcomes, which we use to assess statistical support for candidate improvement.



Specifically, within each bootstrap draw \(j=1,\ldots,J\), we consider
each screening sample \(x\) and harness \(s\in\{-,+\}\) separately.
We draw Dirichlet weights over its \(R\) observed executions,
$
(w_{x,1}^{s,(j)},\ldots,w_{x,R}^{s,(j)})
\sim \operatorname{Dirichlet}(\mathbf 1_R),
$
with $\sum_{r=1}^{R} w_{x,r}^{s,(j)} = 1$, and use the same weight for the quality and cost of each execution.
This produces a plausible reweighted outcome for each screening sample.
Averaging these outcomes equally across screening samples gives
\[
Q_t^{s,(j)}
=
\frac{1}{|\mathcal D_t^{\mathrm{scr}}|}
\sum_{x\in\mathcal D_t^{\mathrm{scr}}}\sum_{r=1}^{R}
w_{x,r}^{s,(j)}q(\tau_{x,r}^{h_t^s}),
\qquad
C_t^{s,(j)}
=
\frac{1}{|\mathcal D_t^{\mathrm{scr}}|}
\sum_{x\in\mathcal D_t^{\mathrm{scr}}}\sum_{r=1}^{R}
w_{x,r}^{s,(j)}c(\tau_{x,r}^{h_t^s}).
\]

We then characterize the candidate's movement in this draw by
\begin{equation}
\Delta Q_t^{(j)}
=
Q_t^{+,(j)}-Q_t^{-,(j)},
\qquad
\Delta C_t^{(j)}
=
\frac{C_t^{+,(j)}-C_t^{-,(j)}}{C_t^{-,(j)}},
\label{eq:method-bootstrap-movement}
\end{equation}
where positive \(\Delta Q_t^{(j)}\) denotes a quality gain and negative
\(\Delta C_t^{(j)}\) denotes a relative cost reduction.

Rather than enforcing a fixed deployment trade-off, we allow improvement in either objective reflecting our key principle: \textit{promote intermediate updates with credible progress along either objective, even if it is not yet desirable for final deployment}. Accordingly, draw \(j\) supports promotion when
\begin{equation}
E_t^{(j)}
=
\bigl[\Delta Q_t^{(j)}>0 \land \Delta C_t^{(j)}\leq\epsilon_C\bigr]
\lor
\bigl[\Delta C_t^{(j)}<0 \land \Delta Q_t^{(j)}\geq-\epsilon_Q\bigr].
\label{eq:method-promotion-event}
\end{equation}
The allowances \(\epsilon_Q,\epsilon_C\geq 0\) serve as loose local safeguards against severe regression in the non-improving objective that may be difficult to recover through subsequent evolution.

Finally, we aggregate the bootstrap outcomes into a promotion-support score:
\begin{equation}
\widehat p_t
=
\frac{1}{J}\sum_{j=1}^{J}\mathbf 1\!\left\{E_t^{(j)}\right\},
\qquad
\text{promote }h'_t
\Longleftrightarrow
\widehat p_t\geq\gamma.
\label{eq:method-promotion-support}
\end{equation}
Here, \(\widehat p_t\) is the fraction of bootstrap draws in which the candidate satisfies the promotion event \(E_t^{(j)}\). Thus, a candidate is promoted only when it is supported by a \(\gamma\) fraction of the resampled outcomes.


\paragraph{Development Evaluation and Candidate Retention.}
Promoted candidates are evaluated on \(\mathcal D_{\mathrm{dev}}\), using
\(R\) executions per task as well. We maintain the set
\(\mathcal V_t\) of all development-evaluated harnesses and its empirical
Pareto archive \(\mathcal A_t\). After evaluating a promoted candidate,
\begin{equation}
\begin{aligned}
\mathcal V_{t+1} = \mathcal V_t\cup\{h'_t\}, \qquad
\mathcal A_{t+1} =
\{h\in\mathcal V_{t+1}:
\nexists g\in\mathcal V_{t+1}\text{ such that }
g\succ_{\mathrm{dev}}h\},
\end{aligned}
\label{eq:method-archive-update}
\end{equation}
where \(\succ_{\mathrm{dev}}\) denotes dominance under the development
estimates. If \(h'_t\) is not promoted, both sets remain unchanged.
Importantly, \(\mathcal V_t\) retains all development-evaluated candidates,
including those outside the current Pareto archive, preserving intermediate
alternatives for subsequent optimization.

\paragraph{Continued Optimization.}
The development-evaluated set \(\mathcal V_{t+1}\), Pareto archive
\(\mathcal A_{t+1}\), accumulated optimization state, and deployment
preference \(\pi\) then guide construction of the next parent \(h_{t+1}\).
Here, \(\pi\) may steer the search toward deployment-relevant regions of the
quality--cost space, but does not act as a hard filter: candidates are not
excluded from further refinement solely because they violate a deployment
constraint or fall outside the current archive.

\paragraph{Deployment Selection.}
After the optimization budget is exhausted, \(\pi\) is applied definitively
to the final development-evaluated set \(\mathcal V\) according to the
weighted, constrained, or Pareto-set formulation in
Section~\ref{sec:formulation}. This yields
\(\mathcal H_\pi^{\mathrm{sel}}\): a single harness for scalar or constrained
selection, or the empirical Pareto set for frontier discovery. 
Thus, deployment preferences \(\pi\) can guide the search without acting as a hard intermediate filter, while governing the final returned solution.

\section{Experiments}

\subsection{Experimental Setup}
\label{sec:exp-setup}

\paragraph{Benchmarks, Agents, and Data Splits.}
We evaluate \framework on two challenging long-horizon agent benchmarks: ClawMark~\citep{meng2026clawmark} and GAIA2~\citep{froger2026gaia2}. For both benchmarks, we construct domain-disjoint training, development, and test splits, such that harness updates are evaluated on previously unseen domains. 
For ClawMark, we use 92 tasks with stage-0 rubrics and optimize OpenClaw~\citep{openclaw2026} with GPT-5.4 at high reasoning effort. For GAIA2, we follow the same split configuration as AutoSaddler~\citep{park2026autosaddler} and optimize the benchmark's default ReAct-style harness using GPT-5.5 with medium reasoning effort.


\paragraph{Instantiation and Baselines.}
We instantiate \framework on top of AutoSaddler.
For deployment preferences, we study two representative settings:
\textbf{(i) Pareto-set optimization}, where no specific quality--cost preference
is specified in advance, and \textbf{(ii) quality-constrained cost minimization},
where cost is minimized subject to a required quality level.

We compare \framework{} against two baselines. \textbf{(i) Meta-Harness}~\citep{lee2026metaharness} is a representative harness-optimization method without staged evaluation, iteratively updating the harness using the full training set. We adapt MH to the two deployment preferences considered above.
\textbf{(ii) AutoSaddler}~\citep{park2026autosaddler} serves as our single-objective, quality-oriented baseline, allowing us to assess whether jointly optimizing quality and cost comes at the expense of quality improvement. 
For conciseness, we denote the evaluated method--preference combinations as
\textsc{PS-Pareto} and \textsc{PS-MinCost} for \framework under Pareto-set
optimization and quality-constrained cost minimization, respectively;
\textsc{MH-Pareto} and \textsc{MH-MinCost} for the corresponding Meta-Harness
variants; and \textsc{AS-Quality} for our quality-only AutoSaddler
adaptation. For Pareto-set optimization, we select two representative harnesses
from the development Pareto set for test evaluation: the quality-extreme
harness (e.g., \textsc{PS-Pareto}$^{Q}$) and the cost-extreme harness (e.g., \textsc{PS-Pareto}$^{C}$).

\paragraph{Experimental Protocol.}
We use the GitHub Copilot SDK~\citep{githubcopilotsdk} with GPT-5.5 at
xhigh reasoning effort as the optimizer for all methods (usage reported in Appendix~\ref{app:optimization_efficiency}). Unless otherwise stated, each task is evaluated with $R=3$ rollouts, and ParetoSaddler
and AutoSaddler use training mini-batches of three tasks. For Bayesian screening, we use \(J=10{,}000\) bootstrap draws, a prespecified promotion threshold of \(\gamma=0.7\), and a local quality-loss allowance of \(\epsilon_Q=0.1\). 
We impose no upper bound on local cost increases, setting \(\epsilon_C=+\infty\). Moreover, for \textsc{PS-MinCost} we require the quality to remain within 10 percentage points below that of the initial harness. We measure response quality using normalized stage-0 rubric score on ClawMark, and Pass@1 on GAIA2, and report token-based execution cost in USD per task. Full experimental details are provided in Appendix~\ref{app:exp_details}.

\begin{table}[!t]
\caption{
Test results on ClawMark and GAIA2.
Quality is reported in percentage (\%), and cost in USD per task.
Values are mean $\pm$ sample standard deviation across three test repeats.
\textbf{Bold} and \underline{underlined} means indicate the best and
second-best results in each column, respectively.
GAIA2 universe-level results are also shown as a detailed breakdown.
For Pareto-set optimization, superscripts \(Q\) and \(C\) denote the
quality- and cost-extreme harnesses selected on the development set.
}
\label{tab:combined_test}
\centering

\resizebox{\textwidth}{!}{%
\begin{tabular}{
@{}lc
GG
@{\hspace{9pt}}
DDD@{}
}
\toprule
\multirow{2}{*}{Setting}
& \multirow{2}{*}{Metric}
& \multicolumn{2}{>{\columncolor{gray!10}}c}{\textbf{Overall}}
& \multicolumn{3}{c}{\footnotesize\textbf{GAIA2 by Universe}} \\
\cmidrule(lr){3-4}
\cmidrule(lr){5-7}
&
& \textbf{ClawMark}
& \textbf{GAIA2}
& U21 (107)
& U22 (112)
& U27 (81) \\
\midrule

\multirow{2}{*}{Default Agent}
& Qual. $\uparrow$
& 70.86 $\pm$ 2.29
& 56.22 $\pm$ 1.07
& 58.26 $\pm$ 1.08
& 53.57 $\pm$ 1.55
& 57.20 $\pm$ 2.57 \\
& Cost $\downarrow$
& 1.17 $\pm$ 0.17
& 0.78 $\pm$ 0.08
& 0.82 $\pm$ 0.18
& 0.71 $\pm$ 0.03
& 0.82 $\pm$ 0.04 \\

\addlinespace[2pt]

\multirow{2}{*}{\textsc{AS-Quality}}
& Qual. $\uparrow$
& 73.86 $\pm$ 3.54
& 57.33 $\pm$ 0.67
& 55.76 $\pm$ 4.70
& 53.87 $\pm$ 5.67
& 64.20 $\pm$ 1.23 \\
& Cost $\downarrow$
& 1.24 $\pm$ 0.04
& 0.61 $\pm$ 0.03
& 0.55 $\pm$ 0.02
& 0.62 $\pm$ 0.04
& 0.67 $\pm$ 0.05 \\

\multirow{2}{*}{\textsc{MH-MinCost}}
& Qual. $\uparrow$
& 70.02 $\pm$ 2.35
& 39.78 $\pm$ 0.96
& 41.12 $\pm$ 2.80
& 38.69 $\pm$ 2.58
& 39.51 $\pm$ 2.47 \\
& Cost $\downarrow$
& 1.16 $\pm$ 0.05
& 0.62 $\pm$ 0.03
& 0.64 $\pm$ 0.04
& 0.57 $\pm$ 0.01
& 0.65 $\pm$ 0.02 \\

\multirow{2}{*}{\textsc{MH-Pareto}$^{C}$}
& Qual. $\uparrow$
& 70.50 $\pm$ 3.92
& 55.44 $\pm$ 1.50
& 58.26 $\pm$ 3.28
& 50.00 $\pm$ 2.68
& 59.26 $\pm$ 3.27 \\
& Cost $\downarrow$
& 0.97 $\pm$ 0.05
& 0.80 $\pm$ 0.01
& 0.76 $\pm$ 0.01
& 0.78 $\pm$ 0.02
& 0.89 $\pm$ 0.02 \\

\multirow{2}{*}{\textsc{MH-Pareto}$^{Q}$}
& Qual. $\uparrow$
& 73.25 $\pm$ 5.88
& 54.33 $\pm$ 2.03
& 57.63 $\pm$ 3.28
& 49.40 $\pm$ 1.36
& 56.79 $\pm$ 3.27 \\
& Cost $\downarrow$
& 1.06 $\pm$ 0.06
& 0.83 $\pm$ 0.04
& 0.87 $\pm$ 0.06
& 0.75 $\pm$ 0.01
& 0.89 $\pm$ 0.07 \\

\midrule

\multirow{2}{*}{\textsc{PS-MinCost}}
& Qual. $\uparrow$
& 75.16 $\pm$ 2.44
& 56.11 $\pm$ 1.71
& 58.26 $\pm$ 0.54
& 50.30 $\pm$ 2.25
& 61.32 $\pm$ 3.77 \\
& Cost $\downarrow$
& \textbf{0.82} $\pm$ 0.07
& \textbf{0.44} $\pm$ 0.01
& \textbf{0.37} $\pm$ 0.02
& \textbf{0.48} $\pm$ 0.01
& 0.48 $\pm$ 0.01 \\

\multirow{2}{*}{\textsc{PS-Pareto}$^{C}$}
& Qual. $\uparrow$
& \underline{79.60} $\pm$ 2.43
& \underline{62.11} $\pm$ 2.46
& \textbf{63.24} $\pm$ 4.80
& \underline{59.23} $\pm$ 1.03
& \underline{64.61} $\pm$ 3.77 \\
& Cost $\downarrow$
& \underline{0.92} $\pm$ 0.08
& 0.47 $\pm$ 0.00
& 0.44 $\pm$ 0.01
& \underline{0.49} $\pm$ 0.02
& \underline{0.47} $\pm$ 0.00 \\

\multirow{2}{*}{\textsc{PS-Pareto}$^{Q}$}
& Qual. $\uparrow$
& \textbf{81.35} $\pm$ 3.11
& \textbf{62.89} $\pm$ 2.14
& \underline{61.99} $\pm$ 2.35
& \textbf{60.71} $\pm$ 2.36
& \textbf{67.08} $\pm$ 1.89 \\
& Cost $\downarrow$
& 1.01 $\pm$ 0.02
& \underline{0.45} $\pm$ 0.02
& \underline{0.39} $\pm$ 0.01
& \underline{0.49} $\pm$ 0.03
& \textbf{0.46} $\pm$ 0.02 \\

\bottomrule
\end{tabular}
}
\vspace{-15pt}
\end{table}

\subsection{Main Results}

Table~\ref{tab:combined_test} reports the performance of
development-selected harnesses on the domain-disjoint test sets, while
Figure~\ref{fig:dev_pareto} shows their evolution in quality--cost space on
the development set. 

\textbf{\framework improves both quality and cost on unseen domains.}
The quality-extreme \textsc{PS-Pareto}$^{Q}$ achieves the highest mean test
quality on both benchmarks while reducing cost relative to the initial
harness. On ClawMark, it improves quality by
\textbf{14.8\%} (70.86\% to 81.35\%) while reducing cost by
\textbf{13.7\%} (\$1.17 to \$1.01); on GAIA2, it improves quality by \textbf{11.9\%}
(56.22\% to 62.89\%) while reducing cost by \textbf{42.3\%}
(\$0.78 to \$0.45 per task).
The cost-extreme \textsc{PS-Pareto}$^{C}$ also improves quality and reduces cost on both benchmarks.
Both operating points outperform their corresponding
Meta-Harness variants in mean quality and cost on both benchmarks.
These gains are consistent across all three held-out GAIA2 universes (see Appendix~\ref{app:stability} for additional stability analysis).

\textbf{Joint optimization does not sacrifice quality, and may even improve it.}
Despite optimizing both quality and cost, \textsc{PS-Pareto}$^{Q}$
outperforms the quality-only \textsc{AS-Quality} baseline by 7.49 quality
points on ClawMark and 5.56 points on GAIA2, while also reducing cost on
both benchmarks. This is a somewhat unexpected result: introducing the cost
objective is associated with better, rather than merely preserved, quality.
We examine this effect and the role of the local Pareto gate in RQ1.

\textbf{Deployment preference steers optimization toward different operating
regions.}
Under quality-constrained cost minimization, \textsc{PS-MinCost} moves
further toward lower-cost solutions while maintaining the required quality level. It reaches \$0.82 per task on ClawMark and \$0.44 on GAIA2, lower than both development-selected \textsc{PS-Pareto} operating points. 
It also dominates
\textsc{MH-MinCost} on both benchmarks in both quality and cost.


\textbf{Pareto reasoning preserves useful quality--cost alternatives.}
As shown in Figure~\ref{fig:dev_pareto}, \textsc{PS-Pareto} explores and
retains multiple operating points along the development quality--cost
frontier rather than committing the search to a single trade-off.
This enables distinct quality- and cost-oriented harnesses to be selected
after optimization according to the deployment preference.

\subsection{Analysis and Ablation Studies}
\label{sec:analysis}


\paragraph{RQ1: How does Pareto-aware promotion enable better optimization outcomes?}
\begin{wrapfigure}[16]{r}{0.43\textwidth}
    \centering
    \vspace{-40pt}
    \includegraphics[
        width=\linewidth,
        trim=12pt 10pt 0 2pt,
        clip
    ]{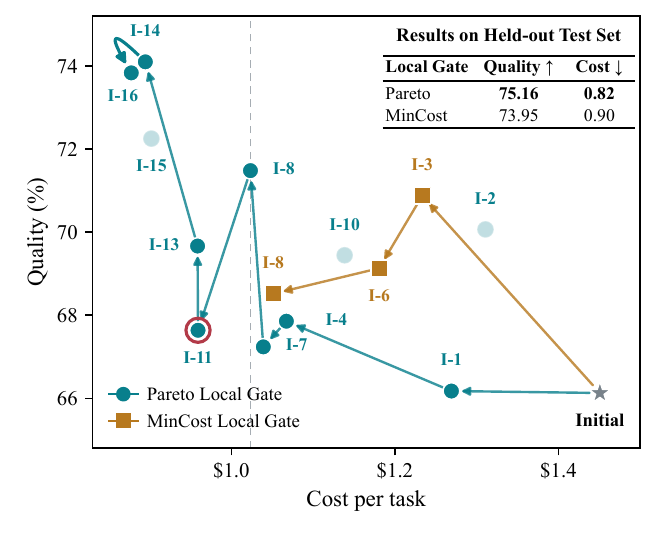}
    \caption{
    ClawMark optimization trajectories and final test results of \textsc{PS-MinCost} under Pareto and
    MinCost local gates.
    Faint markers denote evaluated candidates not selected for subsequent
    refinement.
    }
    \label{fig:gate_ablation}
\end{wrapfigure}
We examine the promotion decisions on ClawMark, with complementary GAIA2 cases in
Appendix~\ref{app:rq1}.

\textbf{Pareto-set vs.\ quality-only optimization.}
I-7 in \textsc{PS-Pareto} provides a representative example (Figure~\ref{fig:clawmark-trajectory}). Its parent
achieves perfect quality on all three mini-batch tasks across three executions. 
I-7 reduces mean mini-batch cost by 17.22\% with a 1.33-point quality decrease, which the quality gate would reject. Pareto-aware promotion instead advances I-7 to development evaluation, where it improves both quality (72.94\% vs.\ 72.27\%) and cost (\$0.88 vs.\ \$1.13).
Its bounded-probing and completion-control behaviors persist in later iterations, contributing to the selected harness. 


\textbf{Controlled gate ablation.}
Because the preceding analysis is observational, we next perform a controlled
study by replacing Pareto-aware promotion with a deployment-oriented
\emph{MinCost Local Gate} in the \textsc{PS-MinCost} setting.
Figure~\ref{fig:gate_ablation} shows that this change produces markedly
different search trajectories. The Pareto gate promotes 11 candidates, compared with only three under the
MinCost Local Gate, and reaches a better test operating point.
The key divergence occurs around I-11, before which the Pareto
trajectory approaches a similar cost ``wall'' to the MinCost Local Gate,
near the dashed line in Figure~\ref{fig:gate_ablation}. 
I-11 improves mini-batch quality from 83.65\% to 87.62\% while increasing cost by 10.5\%.
The MinCost gate would reject this update, whereas Pareto promotion
retains it for development evaluation and subsequent refinement
toward I-14 and I-16.
Without admitting this intermediate update, the Pareto trajectory would likely remain trapped
near the same local cost minima as the MinCost Local Gate. 

Together, these results show that Pareto-aware promotion enables better optimization by preserving intermediate states whose local trade-offs may appear suboptimal, but whose value emerges through broader evaluation and subsequent refinement.

\paragraph{RQ2: How do repeated executions improve harness optimization?}
We ask what additional evidence repeated executions provide and whether this evidence improves optimization outcomes.

\textbf{Diagnostic value of repeated executions.} Repeated runs reveal three types of evidence that a single trajectory cannot provide: \textit{(i) persistent failures}, where repeated failure across runs indicates recurring defects; \textit{(ii) mixed outcomes}, where quality variation on the same task provides direct contrast for unstable behaviors; and \textit{(iii) efficiency contrast}, where all runs succeed but differ substantially in execution cost. Table~\ref{tab:repeated-evidence} provides more details. Repeated executions also strengthen \emph{patch verification and promotion}: pre- and post-patch comparisons reveal whether an improvement persists, while repeated observations provide more reliable evidence for promotion than a single atypical execution.

\begin{wraptable}{r}{0.39\textwidth}
\vspace{-8pt}
\centering
\caption{
Overall ClawMark test results for the single- versus three-run
\textsc{PS-Pareto} ablation. 
}
\label{tab:repeat-ablation-test}

\small
\setlength{\tabcolsep}{4pt}
\renewcommand{\arraystretch}{1.08}
\vspace{-5pt}
\begin{tabular}{@{}lcc@{}}
\toprule
Setting
& Qual. $\uparrow$
& Cost $\downarrow$ \\
\midrule

Single-run$^{C}$
& 76.95 $\pm$ 5.50
& 0.95 $\pm$ 0.06 \\

Single-run$^{Q}$
& 75.71 $\pm$ 2.10
& 0.97 $\pm$ 0.05 \\

\addlinespace[2pt]

Three-run$^{C}$
& 79.60 $\pm$ 2.43
& \textbf{0.92} $\pm$ 0.08 \\

Three-run$^{Q}$
& \textbf{81.35} $\pm$ 3.11
& 1.01 $\pm$ 0.02 \\

\bottomrule
\end{tabular}%

\vspace{-8pt}
\end{wraptable}

\textbf{Single-run vs.\ three-run optimization.}
As the analysis above is again observational, we perform a controlled end-to-end ablation within the \textsc{PS-Pareto} setting, varying only the number of executions per task from one to three while keeping the remaining optimization setup fixed. 
Table~\ref{tab:repeat-ablation-test} shows that three-run optimization reaches stronger held-out operating points. At the cost-oriented endpoint, it achieves 79.60\% quality at \$0.92 per task, compared with 76.95\% at \$0.95 for single-run optimization. At the quality-oriented endpoint, held-out quality increases from 75.71\% to 81.35\%. 

\paragraph{RQ3: Does Bayesian screening improve promotion under execution uncertainty?}

While repeated executions provide richer optimization evidence, this evidence must still be consolidated for promotion. A straightforward mean-based gate compares average quality--cost movements, whereas our Bayesian gate also
accounts for the statistical support behind that movement.
\begin{wraptable}{r}{0.46\textwidth}
    \centering
    \vspace{-6pt}
    \caption{
    ClawMark test results with mean-based versus Bayesian screening.
    }
    \label{tab:mean-vs-bayesian}
    \vspace{-4pt}

    \scriptsize
    \setlength{\tabcolsep}{5.5pt}

    \begin{tabular}{llcc}
        \toprule
        \textbf{Setting}
        & \textbf{Metric}
        & \textbf{Mean}
        & \textbf{Bayesian} \\
        \midrule

        \multirow{2}{*}{\textsc{PS-Pareto}$^{C}$}
        & Qual. $\uparrow$
        & 76.86 $\pm$ 5.41
        & \textbf{79.60} $\pm$ 2.43 \\
        & Cost $\downarrow$
        & 1.09 $\pm$ 0.09
        & \textbf{0.92} $\pm$ 0.08 \\

        \addlinespace[2pt]

        \multirow{2}{*}{\textsc{PS-Pareto}$^{Q}$}
        & Qual. $\uparrow$
        & 76.53 $\pm$ 2.12
        & \textbf{81.35} $\pm$ 3.11 \\
        & Cost $\downarrow$
        & 1.15 $\pm$ 0.04
        & \textbf{1.01} $\pm$ 0.02 \\

        \bottomrule
    \end{tabular}

    \vspace{-6pt}
\end{wraptable}
\textbf{Retrospective analysis of promotion decisions.}
We replay mean-based screening on the same 17 candidates generated by
\textsc{PS-Pareto} on ClawMark. The mean-based gate would promote 14
candidates, versus 12 under Bayesian screening.
I-12 illustrates how aggregation can obscure the underlying
quality--cost movement. Its mean mini-batch quality drops by
5.55 points while cost decreases by 7.19\%, satisfying the configured
quality-loss allowance. Yet Table~\ref{tab:replay} shows that this aggregate
trade-off is highly uneven: nearly all of the cost reduction comes from
\textit{Research assistant 4}, where quality also drops sharply, whereas the
other two tasks become more expensive. 
Retrospective development evaluation further shows that I-12 is 2.80 quality points worse and 10.66\% more expensive than its parent. 

\textbf{Controlled ablation: Bayesian vs.\ mean-based screening.}
Since the retrospective analysis cannot capture how promotion decisions
alter subsequent search, we further perform a full ablation within
\textsc{PS-Pareto}, replacing Bayesian screening with the mean-based gate
while keeping the remaining setup fixed. As shown in
Table~\ref{tab:mean-vs-bayesian}, Bayesian screening reaches stronger test
operating points: its cost-oriented solution achieves 79.60\% quality at
\$0.92 per task, versus 76.86\% at \$1.09 for mean-based screening, while
its quality-oriented solution reaches 81.35\% at \$1.01, versus 76.53\% at
\$1.15.

\textbf{Sensitivity to the promotion threshold.}
Sensitivity analysis in Appendix~\ref{app:RQ3} shows that
\(\gamma=0.7\) lies in a locally stable region on both benchmarks, supporting
its use as a default. For other domains, a small-scale grid search on $\gamma$ may be useful if additional compute is
available.

\section{Conclusion}
We introduced \framework, a framework for staged quality--cost
harness optimization that separates intermediate candidate promotion
from final deployment selection. On ClawMark and GAIA2, its
Pareto-selected harnesses achieve higher quality at lower execution
cost than quality-only harness optimization. Analyses and ablations
show that Pareto-aware promotion preserves intermediate candidates
that can enable later improvements, repeated executions provide richer and more reliable evidence for
optimization, and Bayesian screening filters candidates whose apparent quality--cost gains
are weakly supported under execution uncertainty.
These findings highlight the
value of preserving intermediate trade-offs and accounting for
execution uncertainty in harness optimization.
\section{AI Use Statement}
We used generative AI tools to assist with writing and polishing manuscript sections.
We have not used generative AI tools for research ideation, methodology, or experimental design. All research concepts and analyses were developed and conducted by the authors. 
The authors reviewed the AI-assisted text, checked numerical claims against experimental records, and verified cited sources.
We take full responsibility for the final manuscript, including its claims, results, and conclusions.


\section{Privacy and Responsible AI Considerations}

\paragraph{Privacy.}
Our experiments use publicly available benchmarks,
ClawMark and GAIA2, and do not involve the collection of new
real-user data. We do not investigate privacy-preserving mechanisms
or evaluate the privacy implications of deploying optimized
agent harnesses in real-world applications.

\paragraph{Scope and Responsible AI.}
This work focuses on optimizing agent task quality and execution cost
in benchmark environments. We do not evaluate fairness, harmlessness,
safety, or broader societal impacts. Consequently, improvements in
quality and cost should not be interpreted as evidence of safety or
readiness for unrestricted real-world deployment.

\section{Reproducibility Statement}

We have made every effort to support the reproducibility of the results presented in this paper. 
Section~\ref{sec:method} and Appendix~\ref{app:alg} describe the method and provide pseudocode
for the general framework and its AutoSaddler instantiation.
Section~\ref{sec:exp-setup} and Appendix~\ref{app:exp_details} document the benchmark splits,
model configurations, adaptations,
optimization schedules, and other experimental setup details.
Appendix~\ref{app:trajectories} presents annotated optimization
trajectories, while Appendices~\ref{app:clawmark-search-trace}
and~\ref{app:gaia2-search-trace} document complete runs.
Appendix~\ref{app:prompts} provides selected
optimizer prompt excerpts.
Both benchmarks, ClawMark and GAIA2, are publicly available.
\section{Acknowledgement}
We thank Bo Qiao for his generous help in setting up the computing servers used in our experiments. We are also grateful to the creators and maintainers of ClawMark and GAIA2 for making their benchmarks available, and to the authors of AutoSaddler and Meta-Harness for sharing their work, which informed our experimental comparisons.

\clearpage
\bibliography{reference}

@software{langchain-harness-practice,
  title = {Improving Deep Agents with harness engineering},
  author = {LangChain},
  year = {2026},
  howpublished = {\url{https://www.langchain.com/blog/improving-deep-agents-with-harness-engineering}},
  note={Accessed 2026-04-21}
}

@article{zhou2026externalization,
  title={Externalization in LLM Agents: A Unified Review of Memory, Skills, Protocols and Harness Engineering},
  author={Zhou, Chenyu and Chai, Huacan and Chen, Wenteng and Guo, Zihan and Shan, Rong and Song, Yuanyi and Xu, Tianyi and Yang, Yingxuan and Yu, Aofan and Zhang, Weiming and others},
  journal={arXiv preprint arXiv:2604.08224},
  year={2026}
}

@article{rubin1981bayesian,
  title={The bayesian bootstrap},
  author={Rubin, Donald B},
  journal={The annals of statistics},
  pages={130--134},
  year={1981},
  publisher={JSTOR}
}

@inproceedings{agrawal2026gepa,
  title={Gepa: Reflective prompt evolution can outperform reinforcement learning},
  author={Agrawal, Lakshya A and Tan, Shangyin and Soylu, Dilara and Ziems, Noah and Khare, Rishi and Opsahl-Ong, Krista and Singhvi, Arnav and Shandilya, Herumb and Ryan, Michael J and Jiang, Meng and others},
  booktitle={International Conference on Learning Representations},
  volume={2026},
  pages={8479--8565},
  year={2026}
}

@article{yang2026skillopt,
  title={Skillopt: Executive strategy for self-evolving agent skills},
  author={Yang, Yifan and Gong, Ziyang and Huang, Weiquan and Yang, Qihao and Zhou, Ziwei and Huang, Zisu and Li, Yan and Gao, Xuemei and Dai, Qi and Liu, Bei and others},
  journal={arXiv preprint arXiv:2605.23904},
  year={2026}
}

@inproceedings{guo2024evoprompt,
  title={Connecting large language models with evolutionary algorithms yields powerful prompt optimizers},
  author={Guo, Qingyan and Wang, Rui and Guo, Junliang and Li, Bei and Song, Kaitao and Tan, Xu and Liu, Guoqing and Bian, Jiang and Yang, Yujiu},
  booktitle={International Conference on Learning Representations},
  volume={2024},
  pages={34133--34156},
  year={2024}
}

@article{yuksekgonul2024textgrad,
  title={Textgrad: Automatic "differentiation" via text},
  author={Yuksekgonul, Mert and Bianchi, Federico and Boen, Joseph and Liu, Sheng and Huang, Zhi and Guestrin, Carlos and Zou, James},
  journal={arXiv preprint arXiv:2406.07496},
  year={2024}
}

@article{ni2026trace2skill,
  title={Trace2skill: Distill trajectory-local lessons into transferable agent skills},
  author={Ni, Jingwei and Liu, Yihao and Liu, Xinpeng and Sun, Yutao and Zhou, Mengyu and Cheng, Pengyu and Wang, Dexin and Zhao, Erchao and Jiang, Xiaoxi and Jiang, Guanjun},
  journal={arXiv preprint arXiv:2603.25158},
  year={2026}
}

@article{lee2026metaharness,
  title={Meta-harness: End-to-end optimization of model harnesses},
  author={Lee, Yoonho and Nair, Roshen and Zhang, Qizheng and Lee, Kangwook and Khattab, Omar and Finn, Chelsea},
  journal={arXiv preprint arXiv:2603.28052},
  year={2026}
}

@article{meng2026clawmark,
  title={ClawMark: A Living-World Benchmark for Multi-Turn, Multi-Day, Multimodal Coworker Agents},
  author={Meng, Fanqing and Du, Lingxiao and Wu, Zijian and Chen, Guanzheng and Liu, Xiangyan and Liao, Jiaqi and Jiang, Chonghe and Wan, Zhenglin and Gu, Jiawei and Zhou, Pengfei and others},
  journal={arXiv preprint arXiv:2604.23781},
  year={2026}
}

@inproceedings{froger2026gaia2,
  title={Gaia2: Benchmarking LLM agents on dynamic and asynchronous environments},
  author={Froger, Romain and Andrews, Pierre and Bettini, Matteo and Budhiraja, Amar and Cabral, Ricardo and Do, Virginie and Garreau, Emilien and Gaya, Jean-Baptiste and Lauren{\c{c}}on, Hugo and Lecanu, Maxime and others},
  booktitle={International Conference on Learning Representations},
  volume={2026},
  pages={119758--119789},
  year={2026}
}

@misc{openclaw2026,
  author       = {{OpenClaw}},
  title        = {OpenClaw: Your Own Personal AI Assistant},
  year         = {2026},
  howpublished = {\url{https://github.com/openclaw/openclaw}},
  note         = {GitHub repository. Accessed: 2026-08-25},
}

@article{gong2026skillmoo,
  title={Skillmoo: Multi-objective optimization of agent skills for software engineering},
  author={Gong, Jingzhi and Gu, Ruizhen and Fei, Zhiwei and Cao, Yazhuo and Twist, Lukas and Geiger, Alina and Han, Shuo and Sobania, Dominik and Sarro, Federica and Zhang, Jie M},
  journal={arXiv preprint arXiv:2604.09297},
  year={2026}
}

@article{bussing2026mocapo,
  title={MO-CAPO: Multi-Objective Cost-Aware Prompt Optimization},
  author={B{\"u}ssing, Jan and Schlager, Moritz and Hei{\ss}, Timo and Zehle, Tom and Feurer, Matthias},
  journal={arXiv preprint arXiv:2605.18869},
  year={2026}
}

@inproceedings{yang2024opro,
  title={Large language models as optimizers},
  author={Yang, Chengrun and Wang, Xuezhi and Lu, Yifeng and Liu, Hanxiao and Le, Quoc V and Zhou, Denny and Chen, Xinyun},
  booktitle={International Conference on Learning Representations},
  volume={2024},
  pages={12028--12068},
  year={2024}
}

@article{lou2026autoharness,
  title={Autoharness: improving llm agents by automatically synthesizing a code harness},
  author={Lou, Xinghua and L{\'a}zaro-Gredilla, Miguel and Dedieu, Antoine and Wendelken, Carter and Lehrach, Wolfgang and Murphy, Kevin P},
  journal={arXiv preprint arXiv:2603.03329},
  year={2026}
}

@article{park2026autosaddler,
  title={AutoSaddler: Automatic Harness Optimization with Durable Updates from Agent Execution Traces},
  author={Park, Sungho and Kim, Wonjoong and Tan, Rongyuan and Zhang, Jue and Han, Wook-Shin and Gao, Pengfei and Park, Chanyoung and Yao, Yongqiang and Fu, Rao and Nallipogu, Elsie and others},
  journal={arXiv preprint arXiv:2608.23041},
  year={2026}
}

@article{pan2026rho,
  title={Evolving Agents in the Dark: Retrospective Harness Optimization via Self-Preference},
  author={Pan, Wenbo and Liu, Shujie and Lin, Chin-Yew and Zeng, Jingying and Tang, Xianfeng and Zhou, Xiangyang and Lu, Yan and Jia, Xiaohua},
  journal={arXiv preprint arXiv:2606.05922},
  year={2026}
}

@article{knowles2000paes,
  title={Approximating the nondominated front using the Pareto archived evolution strategy},
  author={Knowles, Joshua D and Corne, David W},
  journal={Evolutionary computation},
  volume={8},
  number={2},
  pages={149--172},
  year={2000},
  publisher={MIT Press}
}

@article{zitzler2001spea2,
  title={SPEA2: Improving the strength Pareto evolutionary algorithm},
  author={Zitzler, Eckart and Laumanns, Marco and Thiele, Lothar},
  journal={TIK report},
  volume={103},
  year={2001},
  publisher={ETH Zurich}
}

@inproceedings{yang2023instoptima,
  title={InstOptima: Evolutionary multi-objective instruction optimization via large language model-based instruction operators},
  author={Yang, Heng and Li, Ke},
  booktitle={Findings of the Association for Computational Linguistics: EMNLP 2023},
  pages={13593--13602},
  year={2023}
}

@article{alzubi2026evoskill,
  title={Evoskill: Automated skill discovery for multi-agent systems},
  author={Alzubi, Salaheddin and Provenzano, Noah and Bingham, Jaydon and Chen, Weiyuan and Vu, Tu},
  journal={arXiv preprint arXiv:2603.02766},
  year={2026}
}

@inproceedings{sinha2024survival,
  title={Survival of the Safest: Towards Secure Prompt Optimization through Interleaved Multi-Objective Evolution},
  author={Sinha, Ankita and Cui, Wendi and Das, Kamalika and Zhang, Jiaxin},
  booktitle={Proceedings of the 2024 Conference on Empirical Methods in Natural Language Processing: Industry Track},
  pages={1016--1027},
  year={2024}
}

@misc{githubcopilotsdk,
  author       = {{GitHub}},
  title        = {GitHub Copilot SDK},
  year         = {2026},
  howpublished = {\url{https://docs.github.com/en/copilot/how-tos/copilot-sdk}},
  note         = {Accessed: 2026-09-08}
}

@article{jiang2026harnessevolve,
  title={HarnessEvolve: Learning from Reference Trajectories for Reliable Agent Self-Evolution},
  author={Jiang, Wen and Chu, Mingmin and Tian, Yimeng and Zhang, Qianxin and Yang, Haofei and Yang, Rui and Liu, Yang and Lv, Tao and Li, Fangming},
  journal={arXiv preprint arXiv:2609.00829},
  year={2026}
}

@inproceedings{pryzant2023apo,
  title={Automatic prompt optimization with “gradient descent” and beam search},
  author={Pryzant, Reid and Iter, Dan and Li, Jerry and Lee, Yin and Zhu, Chenguang and Zeng, Michael},
  booktitle={Proceedings of the 2023 conference on empirical methods in natural language processing},
  pages={7957--7968},
  year={2023}
}

@InProceedings{knowles2001multiobjectivization,
author="Knowles, Joshua D.
and Watson, Richard A.
and Corne, David W.",
editor="Zitzler, Eckart
and Thiele, Lothar
and Deb, Kalyanmoy
and Coello Coello, Carlos Artemio
and Corne, David",
title="Reducing Local Optima in Single-Objective Problems by Multi-objectivization",
booktitle="Evolutionary Multi-Criterion Optimization",
year="2001",
publisher="Springer Berlin Heidelberg",
address="Berlin, Heidelberg",
pages="269--283",
isbn="978-3-540-44719-1"
}

@article{lehman2011novelty,
    author = {Lehman, Joel and Stanley, Kenneth O.},
    title = {Abandoning Objectives: Evolution Through the Search for Novelty Alone},
    journal = {Evolutionary Computation},
    volume = {19},
    number = {2},
    pages = {189-223},
    year = {2011},
    month = {06},
    issn = {1063-6560},
    doi = {10.1162/EVCO_a_00025},
    url = {https://doi.org/10.1162/EVCO_a_00025},
    eprint = {https://direct.mit.edu/evco/article-pdf/19/2/189/1494066/evco_a_00025.pdf},
}

@misc{schmucker2021multiobjectiveasynchronoussuccessivehalving,
      title={Multi-objective Asynchronous Successive Halving}, 
      author={Robin Schmucker and Michele Donini and Muhammad Bilal Zafar and David Salinas and Cédric Archambeau},
      year={2021},
      eprint={2106.12639},
      archivePrefix={arXiv},
      primaryClass={stat.ML},
      url={https://arxiv.org/abs/2106.12639}, 
}

@misc{zhang2026hyperagents,
      title={Hyperagents}, 
      author={Jenny Zhang and Bingchen Zhao and Wannan Yang and Jakob Foerster and Jeff Clune and Minqi Jiang and Sam Devlin and Tatiana Shavrina},
      year={2026},
      eprint={2603.19461},
      archivePrefix={arXiv},
      primaryClass={cs.AI},
      url={https://arxiv.org/abs/2603.19461}, 
}

@InProceedings{branke2003selection,
author="Branke, J{\"u}rgen
and Schmidt, Christian",
editor="Cant{\'u}-Paz, Erick
and Foster, James A.
and Deb, Kalyanmoy
and Davis, Lawrence David
and Roy, Rajkumar
and O'Reilly, Una-May
and Beyer, Hans-Georg
and Standish, Russell
and Kendall, Graham
and Wilson, Stewart
and Harman, Mark
and Wegener, Joachim
and Dasgupta, Dipankar
and Potter, Mitch A.
and Schultz, Alan C.
and Dowsland, Kathryn A.
and Jonoska, Natasha
and Miller, Julian",
title="Selection in the Presence of Noise",
booktitle="Genetic and Evolutionary Computation --- GECCO 2003",
year="2003",
publisher="Springer Berlin Heidelberg",
address="Berlin, Heidelberg",
pages="766--777",
isbn="978-3-540-45105-1"
}

@InProceedings{marceau2014racing,
author="Marceau-Caron, Gaetan
and Schoenauer, Marc",
editor="Bartz-Beielstein, Thomas
and Branke, J{\"u}rgen
and Filipi{\v{c}}, Bogdan
and Smith, Jim",
title="Racing Multi-objective Selection Probabilities",
booktitle="Parallel Problem Solving from Nature -- PPSN XIII",
year="2014",
publisher="Springer International Publishing",
address="Cham",
pages="631--640",
isbn="978-3-319-10762-2"
}

@inproceedings{rook2024multiobjective,
author = {Rook, Jeroen and Hoos, Holger H. and Trautmann, Heike},
title = {Multi-objective Ranking using Bootstrap Resampling},
year = {2024},
isbn = {9798400704956},
publisher = {Association for Computing Machinery},
address = {New York, NY, USA},
url = {https://doi.org/10.1145/3638530.3654436},
doi = {10.1145/3638530.3654436},
booktitle = {Proceedings of the Genetic and Evolutionary Computation Conference Companion},
pages = {155–158},
numpages = {4},
location = {Melbourne, VIC, Australia},
series = {GECCO '24 Companion}
}

@inproceedings{vattikonda2025train,
 author = {Vattikonda, Dheeraj and Ravichandran, Santhoshi and Penaloza, Emiliano and Nekoei, Hadi and de Chezelles, Thibault and Thakkar, Megh and Gontier, Nicolas and Mu\~{n}oz-M\'{a}rmol, Miguel and Omidi Shayegan, Sahar and Raimondo, Stefania and Liu, Steve (Xue) and Drouin, Alexandre and Piche, Alexandre and Lacoste, Alexandre and Caccia, Massimo},
 booktitle = {Advances in Neural Information Processing Systems},
 doi = {10.52202/085713-1774},
 editor = {D. Belgrave and C. Zhang and H. Lin and R. Pascanu and P. Koniusz and M. Ghassemi and N. Chen},
 pages = {53256--53282},
 publisher = {Curran Associates, Inc.},
 title = {How to Train Your LLM Web Agent: A Statistical Diagnosis},
 url = {https://proceedings.neurips.cc/paper_files/paper/2025/file/4d239452d41ded07139801d7c966c629-Paper-Conference.pdf},
 volume = {38, Main Conference},
 year = {2025}
}
\bibliographystyle{iclr2027_conference}

\clearpage
\appendix
\section{Additional Experimental Details}
\label{app:exp_details}

\textbf{Benchmarks and Data Splits.}
We evaluate \framework on two long-horizon agent benchmarks with complementary execution settings. ClawMark~\citep{meng2026clawmark} evaluates multi-turn, multi-day coworker agents across stateful professional scenarios, covering realistic workflows that involve persistent state, tool use, and rule-based evaluation. GAIA2~\citep{froger2026gaia2} evaluates agents in dynamic and asynchronous environments, where the environment may evolve independently of agent actions and tasks require sustained interaction with external services.

We construct training, development, and test splits at the group level rather than randomly partitioning individual tasks. This prevents closely related tasks from the same group from appearing in both optimization and evaluation, and tests whether harness updates learned during optimization transfer to previously unseen task groups. The same splits are fixed before optimization and shared by all
methods. For ClawMark, we restrict evaluation to initial-stage tasks with explicit stage-0 rubrics to keep repeated evaluations across optimization candidates within the available execution budget. Later-stage adaptation is outside the scope of this evaluation. For GAIA2, we follow the split protocol of AutoSaddler~\citep{park2026autosaddler}. The resulting assignments are summarized in Table~\ref{tab:benchmark_splits}.

\begin{table}[H]
\centering
\caption{
Group-disjoint benchmark splits. ClawMark is partitioned by scenario and
GAIA2 by Universe. Numbers in parentheses denote the number of tasks.
}
\label{tab:benchmark_splits}

\small
\begingroup
\setlength{\tabcolsep}{5pt}
\renewcommand{\arraystretch}{1.15}

\begin{tabular}{
@{}
>{\raggedright\arraybackslash}p{0.12\linewidth}
>{\raggedright\arraybackslash}p{0.31\linewidth}
>{\raggedright\arraybackslash}p{0.23\linewidth}
>{\raggedright\arraybackslash}p{0.25\linewidth}
@{}
}
\toprule
\textbf{Benchmark}
& \textbf{Train}
& \textbf{Dev}
& \textbf{Test} \\
\midrule

\textbf{ClawMark}
& Clinical Assistant, Content Operation, EDA,\newline
  Executive Assistant, HR, Research Assistant (50)
& Investment Analyst, Project Management,\newline
  Real Estate (20)
& E-commerce, Journalist,\newline
  Legal Assistant (22) \\

\addlinespace[3pt]

\textbf{GAIA2}
& Universe 29 (75)
& Universe 30 (65)
& Universes 21, 22, 27 (300) \\

\bottomrule
\end{tabular}
\endgroup
\end{table}

\textbf{Cost Accounting.}
We use GPT-5.4 with high reasoning effort for ClawMark experiments, and GPT-5.5 with medium reasoning effort for GAIA2 experiments.
Table~\ref{tab:model-pricing} reports the token prices used to compute execution cost according to $
c(\tau)=\sum_{j=1}^{M(\tau)}
\left(
\rho_{\mathrm{in}}T_{j,\mathrm{in}}
+\rho_{\mathrm{cache}}T_{j,\mathrm{cache}}
+\rho_{\mathrm{out}}T_{j,\mathrm{out}}
\right)$. We charge fresh-input, cached-input, and output tokens separately using the price card of the corresponding task model.

\begin{table}[!h]
\centering
\caption{Task-model configurations and token prices used for execution-cost
accounting. Prices are in USD per one million tokens.}
\label{tab:model-pricing}
\small
\begin{tabular}{llcccc}
\toprule
\textbf{Benchmark} &
\textbf{Model} &
\textbf{Reasoning} &
\textbf{Fresh input} &
\textbf{Cached input} &
\textbf{Output} \\
\midrule
ClawMark & GPT-5.4 & High   & \$2.50 & \$0.25 & \$15.00 \\
GAIA2    & GPT-5.5 & Medium & \$5.00 & \$0.50 & \$30.00 \\
\bottomrule
\end{tabular}
\end{table}

We normalize execution cost to a cold-start
accounting convention: cached input tokens reported for the first valid model call in each task execution are charged at the fresh-input rate. Subsequent calls use the reported fresh-input and cache-read token counts at their respective rates. This convention prevents pre-existing cache state from discounting the initial call in the reported execution cost.

\textbf{Deployment Settings.}
We evaluate two settings for deployment preference: Pareto-set optimization and quality-constrained cost minimization. The symmetric setting of cost-constrained quality maximization can be handled analogously. We do not separately evaluate weighted-sum scalarization, as its fixed scalar objective can be optimized by existing
single-objective AHO methods. 

\textbf{Benchmark Adaptation.}
All ClawMark experiments use OpenClaw v2026.6.9. We adapt ClawMark to reduce cross-run interference during parallel
evaluation. Each execution receives a unique Notion root, replacing
shared evaluation-hub cleanup with execution-scoped teardown.
For Google Sheets, we retain per-execution Drive folders and strengthen
exact-ID resource lookup. Execution-specific resource IDs are injected
into each stage's agent context and reused in checker-side resolution,
while tool-use guidance discourages global title-based discovery.
These changes preserve task objectives and rubric criteria.

\textbf{Baseline Adaptation.}
We adapt AutoSaddler and Meta-Harness to the same benchmark environments
and monetary-cost accounting used by \framework, while preserving their
distinct search and selection mechanisms. Because the methods differ in their
evaluation structure, we report their evaluation schedules and repetition
counts explicitly rather than forcing an identical protocol.

For AutoSaddler~\citep{park2026autosaddler}, we use its quality-only
configuration, denoted \textsc{AS-Quality}. We combine its separate capability- and steering-oriented optimization phases into a unified search loop, allowing the
optimizer to apply either or both types of changes within each iteration. This aligns the available edit choices with \framework and avoids confounding the comparison of optimization objectives with a fixed phase schedule. Each task is executed three times, with quality averaged across executions before aggregation across tasks. We also apply a comparable Bayesian-bootstrap test when deciding whether to promote a local candidate based on quality alone, using the same promotion threshold $\gamma$. Proposal generation and candidate selection therefore remain quality-oriented; execution cost is only recorded for evaluation but is not used as an optimization objective or promotion criterion.

For Meta-Harness~\citep{lee2026metaharness}, we retain its
history-conditioned proposal and evaluation loop without introducing staged evaluation. We instantiate two
objective-specific variants. \textsc{MH-Pareto} receives both quality and monetary-cost
feedback, maintains the nondominated quality--cost frontier, and draws proposal
parents from this frontier. \textsc{MH-MinCost} instead minimizes mean execution cost
subject to the baseline-relative quality constraint
\[
Q(h) \geq \max\{0, Q(h_0)-\epsilon\},
\]
using the same absolute tolerance $\epsilon$ as \textsc{PS-MinCost}. The lowest-cost
feasible candidate serves as the selection reference, while other historical candidates still contribute to subsequent proposals.

\textbf{Optimization Details.}
Unless otherwise stated, we use $R=3$ executions per task. This choice reflects the substantial
cost of repeatedly executing long-horizon agents across optimization
candidates and the available evaluation budget. With only three
executions, rare outcomes may remain unobserved, and estimates of
execution variability and bootstrap promotion support may be sensitive
to the observed runs, particularly for candidates near the promotion
threshold. Larger repetition counts could improve statistical precision
but would require additional evaluation resources. We use $J=10{,}000$ independent Bayesian bootstrap draws, limiting the
Monte Carlo standard error of the estimated promotion support
to at most 0.005. These draws reuse the recorded observations and require no additional LLM calls.
The cost safeguard is disabled by setting \(\epsilon_C=+\infty\). This is because a substantial local cost increase can still reflect genuine task progress. For example, resolving premature termination may result in substantial task progress while requiring more reasoning or environment interactions, making the cost increase difficult to estimate beforehand.

ParetoSaddler and \textsc{AS-Quality} perform one optimization epoch over the
training split using minibatches of three tasks. This yields 17 proposal
iterations on ClawMark and 25 on GAIA2. Unless otherwise stated, each task
is executed three times. The task model and its inference
configuration remain fixed throughout optimization.

Meta-Harness uses its original full-set evaluation structure rather than
staged evaluation. Each proposed harness is evaluated on the combined
training and development search set. We run Meta-Harness for 8 proposal iterations on ClawMark and
15 on GAIA2, giving Meta-Harness approximately $1.37\times$ and $1.22\times$ the
maximum task-execution budgets of the staged methods on ClawMark and GAIA2, respectively. We allocate Meta-Harness a larger evaluation budget to avoid favoring the methods using staged evaluation through additional optimization compute.

\begin{table}[!h]
\centering
\caption{
\textbf{Optimization schedules and task-execution budgets.}
A task execution denotes one complete rollout of the task agent on one
benchmark task. For ParetoSaddler (\textsc{PS}) and AutoSaddler (\textsc{AS}), the reported budget is an
upper bound that assumes every locally screened candidate is promoted to
full-development evaluation; the actual number of task executions is therefore
less than or equal to this value and depends on the number of local promotions.
Meta-Harness (\textsc{MH}) does not use staged promotion and instead evaluates each proposal
on the full search set, so its budget is determined directly by the number of
proposals shown.
All budgets include the initial default harness evaluation.
}
\label{tab:optimization_budget}
\small
\begin{tabular}{llcccc}
\toprule
Benchmark &
Method &
Proposals &
Search tasks / proposal &
Repeats &
Task-execution budget \\
\midrule
ClawMark
& \textsc{PS} / \textsc{AS}
& 17
& 3 + promoted dev. eval.
& 3
& 1380 \\
& \textsc{MH}
& 8
& 70
& 3
& 1890 \\
\midrule
GAIA2
& \textsc{PS} / \textsc{AS}
& 25
& 3 + promoted dev. eval.
& 3
& 5520 \\
& \textsc{MH}
& 15
& 140
& 3
& 6720 \\
\bottomrule
\end{tabular}
\end{table}
\section{Algorithm of \framework}
\label{app:alg}

\paragraph{General Algorithm.}
Algorithm~\ref{alg:paretosaddler-general} gives the complete procedure for
iterative AHO with staged evaluation. We use the same notation as
Section~\ref{sec:method}. At iteration \(t\), generation samples
\(\mathcal D_t^{\mathrm{gen}}\subseteq\mathcal D_{\mathrm{train}}\)
provide the evidence used to construct \(h'_t\), while screening samples
\(\mathcal D_t^{\mathrm{scr}}\subseteq\mathcal D_{\mathrm{train}}\)
provide the repeated parent--candidate evidence used for promotion.
The two sample sets may differ or overlap, and compatible parent executions
may be reused.

Let \(h_t^{-}=h_t\) and \(h_t^{+}=h'_t\). For
\(s\in\{-,+\}\), repeated screening produces
\[
\mathcal T_t^s
=
\{\tau_{x,r}^{h_t^s}:
x\in\mathcal D_t^{\mathrm{scr}},\ r=1,\ldots,R\},
\qquad
\mathcal Z_t^s
=
\{\mathbf z_{x,r}^{h_t^s}:
x\in\mathcal D_t^{\mathrm{scr}},\ r=1,\ldots,R\},
\]
where
\[
\mathbf z_{x,r}^{h_t^s}
=
\bigl(q(\tau_{x,r}^{h_t^s}),
c(\tau_{x,r}^{h_t^s})\bigr).
\]

\paragraph{Bootstrap Pareto Support.}
For completeness, we give the exact estimator used by
\textsc{BootstrapParetoSupport}. For each bootstrap draw
\(j=1,\ldots,J\), screening sample
\(x\in\mathcal D_t^{\mathrm{scr}}\), and
\(s\in\{-,+\}\), independently draw
\[
\mathbf w_{t,x}^{s,(j)}
\sim\operatorname{Dirichlet}(\mathbf 1_R),
\qquad
\sum_{r=1}^{R}w_{t,x,r}^{s,(j)}=1.
\]
Applying each weight jointly to the corresponding quality and cost gives
the task-balanced estimates
\[
Q_t^{s,(j)}
=
\frac{1}{|\mathcal D_t^{\mathrm{scr}}|}
\sum_{x\in\mathcal D_t^{\mathrm{scr}}}
\sum_{r=1}^{R}
w_{t,x,r}^{s,(j)}
q(\tau_{x,r}^{h_t^s}),
\qquad
C_t^{s,(j)}
=
\frac{1}{|\mathcal D_t^{\mathrm{scr}}|}
\sum_{x\in\mathcal D_t^{\mathrm{scr}}}
\sum_{r=1}^{R}
w_{t,x,r}^{s,(j)}
c(\tau_{x,r}^{h_t^s}).
\]
We then compute
\[
\Delta Q_t^{(j)}
=
Q_t^{+,(j)}-Q_t^{-,(j)},
\qquad
\Delta C_t^{(j)}
=
\frac{C_t^{+,(j)}-C_t^{-,(j)}}{C_t^{-,(j)}},
\]
and define
\[
E_t^{(j)}
=
\bigl[
\Delta Q_t^{(j)}>0
\land
\Delta C_t^{(j)}\leq\epsilon_C
\bigr]
\lor
\bigl[
\Delta C_t^{(j)}<0
\land
\Delta Q_t^{(j)}\geq-\epsilon_Q
\bigr].
\]
The estimated promotion support is
\[
\widehat p_t
=
\frac{1}{J}
\sum_{j=1}^{J}
\mathbf 1\!\left\{E_t^{(j)}\right\}.
\]
Algorithm~\ref{alg:paretosaddler-general} promotes the candidate when
\(\widehat p_t\geq\gamma\).

\paragraph{AutoSaddler Instantiation.}
Algorithm~\ref{alg:paretosaddler-autosaddler} instantiates \framework on
AutoSaddler~\citep{park2026autosaddler}, with adaptations highlighted in
\textcolor{orange}{orange}. At iteration \(t\), the same training mini-batch
is used for generation and screening,
\[
\mathcal D_t^{\mathrm{gen}}
=
\mathcal D_t^{\mathrm{scr}}
=
\mathcal B_t,
\]
so the repeated parent executions are reused as generation evidence.
AutoSaddler diagnoses task failures and inefficient behaviors from these
executions, proposes a targeted patch, and evaluates the resulting candidate
on the same mini-batch before applying the Pareto promotion gate.
Promoted candidates receive development evaluation; subsequent reflection
and EvoDAG evolution use the accumulated evidence together with
\(\mathcal V_{t+1}\), \(\mathcal A_{t+1}\), and \(\pi\).

\begin{algorithm}[t]
\small
\caption{ParetoSaddler for iterative AHO with staged evaluation}
\label{alg:paretosaddler-general}
\begin{algorithmic}[1]
\Require Initial harness \(h_0\), development set
\(\mathcal D_{\mathrm{dev}}\), optimization budget \(K\),
number of evaluation repeats \(R\), number of bootstrap draws
\(J\), parent-relative local allowances
\(\epsilon_Q,\epsilon_C\), promotion threshold \(\gamma\),
and deployment preference \(\pi\)

\State Evaluate \(h_0\) on \(\mathcal D_{\mathrm{dev}}\)
with \(R\) repeats and record the results

\State Initialize optimization state \(\mathcal S_0\)
with \(h_0\) and its evaluation results

\State Initialize development-evaluated set
\(\mathcal V_0\gets\{h_0\}\) and Pareto archive
\(\mathcal A_0\gets\{h_0\}\)

\For{iteration \(n=0,1,2,\ldots\) until budget \(K\) is exhausted}
    \State Select candidate-generation context
    \(\mathcal C_n^{\mathrm{gen}}\)

    \State \((\mathcal T_n^{\mathrm{gen}},
    \mathcal Z_n^{\mathrm{gen}})
    \gets\Call{CollectEvidence}
    {h_n,\mathcal C_n^{\mathrm{gen}}}\)

    \State \(h'_n
    \gets\Call{GenerateCandidate}
    {h_n,\mathcal T_n^{\mathrm{gen}},
    \mathcal Z_n^{\mathrm{gen}},\mathcal S_n}\)

    \State Select preliminary screening context
    \(\mathcal C_n^{\mathrm{scr}}\)

    \State \((\mathcal T_n^{-},\mathcal Z_n^{-})
    \gets\Call{Evaluate}
    {h_n,\mathcal C_n^{\mathrm{scr}},R}\),
    reusing compatible parent executions when available

    \State \((\mathcal T_n^{+},\mathcal Z_n^{+})
    \gets\Call{Evaluate}
    {h'_n,\mathcal C_n^{\mathrm{scr}},R}\)

    \State \(\widehat p_n
    \gets\Call{BootstrapParetoSupport}
    {\mathcal Z_n^{-},\mathcal Z_n^{+},
    \epsilon_Q,\epsilon_C,J}\)

    \If{\(\widehat p_n\geq\gamma\)}
        \State Evaluate \(h'_n\) on
        \(\mathcal D_{\mathrm{dev}}\) with \(R\) repeats
        and record the results

        \State \(\mathcal V_{n+1}
        \gets\mathcal V_n\cup\{h'_n\}\)

        \State \(\mathcal A_{n+1}
        \gets\Call{UpdateParetoArchive}{\mathcal V_{n+1}}\)
    \Else
        \State \(\mathcal V_{n+1}\gets\mathcal V_n\)
        \State \(\mathcal A_{n+1}\gets\mathcal A_n\)
    \EndIf

    \State \(\mathcal S_{n+1}
    \gets\Call{UpdateState}
    {\mathcal S_n,h_n,h'_n,\text{all available evidence}}\)

    \State \(h_{n+1}
    \gets\Call{ContinueOptimization}
    {\mathcal V_{n+1},\mathcal A_{n+1},\mathcal S_{n+1},\pi}\)
\EndFor

\State \(\mathcal H_\pi^{\mathrm{sel}}
\gets\Call{SelectForDeployment}{\mathcal V,\pi}\)

\State \Return \(\mathcal H_\pi^{\mathrm{sel}}\)
\end{algorithmic}
\end{algorithm}

\begin{algorithm}[t]
\small
\caption{ParetoSaddler instantiated on AutoSaddler
(adaptations highlighted in \textcolor{orange}{orange})}
\label{alg:paretosaddler-autosaddler}
\begin{algorithmic}[1]
\Require Initial harness \(h_0\), training set
\(\mathcal D_{\mathrm{train}}\), development set
\(\mathcal D_{\mathrm{dev}}\), rollout budget \(K\),
\textcolor{orange}{number of evaluation repeats \(R\),
number of bootstrap draws \(J\), parent-relative local
allowances \(\epsilon_Q,\epsilon_C\), promotion threshold
\(\gamma\), and deployment preference \(\pi\)}

\State Initialize EvoDAG \(G_0\) with \(h_0\)

\State Evaluate \(h_0\) on \(\mathcal D_{\mathrm{dev}}\)
\textcolor{orange}{with \(R\) repeats}
and record the results in \(G_0\)

\State Initialize development-evaluated set
\(\mathcal V_0\gets\{h_0\}\)
\textcolor{orange}{and Pareto archive
\(\mathcal A_0\gets\{h_0\}\)}

\For{iteration \(n=0,1,2,\ldots\) until budget \(K\) is exhausted}
    \State Sample mini-batch
    \(\mathcal B_n\subset\mathcal D_{\mathrm{train}}\)

    \State \((\mathcal T_n^{-},\mathcal Z_n^{-})
    \gets\Call{Evaluate}
    {h_n,\mathcal B_n,\textcolor{orange}{R}}\)

    \State Diagnose failures
    \textcolor{orange}{and inefficiencies across repeated executions}
    using \(\mathcal T_n^{-}\) and \(\mathcal Z_n^{-}\)

    \State Generate a targeted patch \(\Delta\theta_n\), yielding
    candidate \(h'_n=h_n+\Delta\theta_n\)

    \State \((\mathcal T_n^{+},\mathcal Z_n^{+})
    \gets\Call{Evaluate}
    {h'_n,\mathcal B_n,\textcolor{orange}{R}}\)

    \State \textcolor{orange}{
    \(\widehat p_n
    \gets\Call{BootstrapParetoSupport}
    {\mathcal Z_n^{-},\mathcal Z_n^{+},
    \epsilon_Q,\epsilon_C,J}\)}

    \If{\(\textcolor{orange}{\widehat p_n\geq\gamma}\)}
        \State Evaluate \(h'_n\) on \(\mathcal D_{\mathrm{dev}}\)
        \textcolor{orange}{with \(R\) repeats}
        and record the results

        \State \(\mathcal V_{n+1}
        \gets\mathcal V_n\cup\{h'_n\}\)

        \State \textcolor{orange}{
        \(\mathcal A_{n+1}
        \gets\Call{UpdateParetoArchive}{\mathcal V_{n+1}}\)}
    \Else
        \State \(\mathcal V_{n+1}\gets\mathcal V_n\)

        \State \textcolor{orange}{
        \(\mathcal A_{n+1}\gets\mathcal A_n\)}
    \EndIf

    \State Reflect on \(\mathcal T_n^{-}\),
    \(\mathcal T_n^{+}\), and available evaluation results

    \State Update EvoDAG \(G_n\) with the candidate, patch,
    metrics, and lessons to obtain \(G_{n+1}\)

    \State Evolve \(h_{n+1}\) from \(G_{n+1}\),
    \textcolor{orange}{guided by \(\mathcal V_{n+1}\),
    \(\mathcal A_{n+1}\), and \(\pi\)}
\EndFor

\State \textcolor{orange}{
\(\mathcal H_\pi^{\mathrm{sel}}
\gets\Call{SelectForDeployment}{\mathcal V,\pi}\)}

\State \Return \textcolor{orange}{\(\mathcal H_\pi^{\mathrm{sel}}\)}
\end{algorithmic}
\end{algorithm}

\section{Additional Analysis for RQ1}
\label{app:rq1}

\paragraph{Local MinCost Gate.}
For the gate comparison in Section~\ref{sec:analysis}, we replace Pareto-aware
promotion with a cost-oriented local screening rule within
\textsc{PS-MinCost}. This alternative retains repeated execution
and Bayesian-bootstrap screening, but requires supported cost
reduction rather than progress in either objective.

Using the same parent-relative quality difference
$\Delta Q_t^{(j)}$ and relative cost change
$\Delta C_t^{(j)}$, the promotion event is
\[
E_{t,\mathrm{MC}}^{(j)}
=
\left\{
\Delta C_t^{(j)} < 0
\;\land\;
\Delta Q_t^{(j)} \geq -\epsilon_Q
\right\}.
\]
The joint support and promotion decision are
\[
\widehat p_{t,\mathrm{MC}}
=
\frac{1}{J}\sum_{j=1}^{J}
\mathbf{1}\!\left\{E_{t,\mathrm{MC}}^{(j)}\right\},
\qquad
\text{promote }h'_t
\iff
\widehat p_{t,\mathrm{MC}} \geq \gamma.
\]
This rule retains only the cost-improvement branch of
Pareto promotion: it permits cost reductions with bounded
quality loss but does not credit quality gains without
cost reduction. 

\paragraph{GAIA2 analysis.}

We complement the ClawMark analysis in Section~\ref{sec:analysis} with case studies from the GAIA2 \textsc{PS-Pareto} run. We focus on I-9 and I-17, whose mini-batches achieve perfect quality. 

Iteration I-9 achieves perfect quality on all three mini-batch tasks across all three executions before applying patches. Nevertheless, the execution traces expose unnecessary retrieval overhead: successful aggregation workflows retain full email bodies and perform individual contact lookups. I-9 then introduces compact, date-filtered email reads and batched contact lookup, reducing the amount of information retrieved and the number of separate lookup operations required. After the update, screening quality remains at 100\%, while mean cost decreases from \$0.824 to \$0.494 per task, by 40.02\%. Since every observed quality score is 100\% for both harnesses, the quality-only gate would reject the promotion because it requires a strict quality improvement. The Pareto gate instead assigns promotion support of 1.0 and advances I-9 to development evaluation. On the development set, I-9 improves both objectives relative to its parent: quality increases from 54.36\% to 55.90\%, while cost decreases from \$0.64 to \$0.54 per task, leading to a better development operating point.

I-17 provides another example involving both evidence access and execution control. Its parent again succeeds in all nine screening executions, but successful trajectories still contain inefficient access to email history and unnecessary waiting after task completion. The update adds recency information to summaries, supports oldest-first email retrieval, and introduces termination control to avoid continued idle waiting when no follow-up work remains. Screening quality stays at perfect quality, while mean cost decreases from \$0.50 to \$0.30 per task, a 39.96\% reduction. A strict quality-only gate will reject the promotion, whereas Pareto promotion support is 1.0. Development evaluation subsequently shows a quality increase from 56.92\% to 60.00\%, alongside a cost reduction from \$0.38 to \$0.37 per task. The benefit observed on the screening tasks therefore extends beyond lower local execution cost to a stronger quality--cost operating point on development tasks.

Together with the ClawMark case, these examples show that local quality saturation does not imply that a harness has exhausted its improvement opportunities. Cost supplies an additional promotion signal, allowing updates to retrieval and execution control to receive development evaluation even when screening quality cannot increase.
\begin{figure}[!h]
   \centering

    \includegraphics[width=\linewidth]{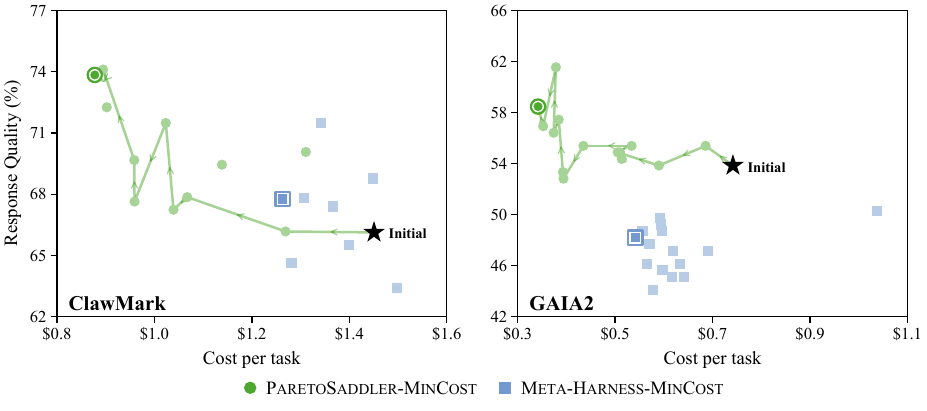}
    \caption{\textbf{Quality--cost outcomes under quality-constrained cost minimization on the development sets of ClawMark and GAIA2.}
    Each marker represents a harness evaluated on the development set;
    better solutions lie toward the upper left.
    Stars denote the initial harness, and enlarged outlined markers
    highlight the final candidates selected by each method.
    Solid green lines trace harness evolution in \textsc{PS-MinCost};
    green markers not connected by solid lines denote candidates
    not selected for further refinement. The selected \textsc{PS-MinCost} harnesses achieve substantially
    lower cost than their \textsc{MH-MinCost} counterparts
    while also attaining higher quality.
    }
    \label{fig:mincost_dev}
\end{figure}
\paragraph{Development-set comparison under cost minimization.}
Figure~\ref{fig:mincost_dev} compares \textsc{PS-MinCost}
and \textsc{MH-MinCost} on the development sets of ClawMark
and GAIA2. On both benchmarks, \textsc{PS-MinCost} identifies
lower-cost harnesses with higher development quality than
the evaluated \textsc{MH-MinCost} candidates.
These results complement the held-out comparisons in
Table~\ref{tab:combined_test}.

\section{Additional Analysis for RQ2}
\label{app:rq2}

\paragraph{Variation of Response Quality and Cost across Executions.}
\label{app:quality_cost_variations}

\begin{figure}[!h]
    \centering
    \includegraphics[width=0.44\linewidth]{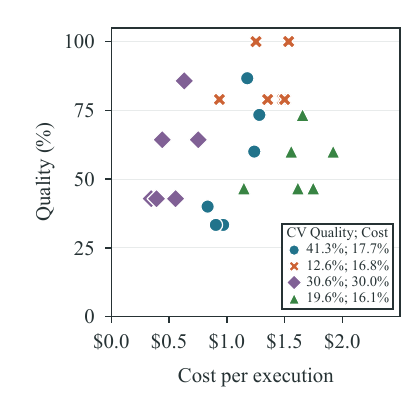}
    \caption{
        Run-to-run variability on four sampled ClawMark tasks.
        Markers show six executions per task; the legend gives
        within-task CV for quality and cost.
    }
    \label{fig:execution-variability}
\end{figure}

Execution randomness can affect both objectives. 
Figure~\ref{fig:execution-variability} illustrates this effect for a fixed
harness evaluated six times on four randomly sampled ClawMark tasks.
Variability is also evident during optimization:
across the 50 training tasks in the \textsc{PS-Pareto} run,
the mean per-task coefficient of variation (CV) is
20.35\% for quality and 23.30\% for cost before patching,
and remains 17.05\% and 21.86\%, respectively, after patching.
Across the 20 development tasks, the corresponding CVs are
18.8\% for quality and 25.6\% for cost.
These observations motivate repeated execution when assessing
candidate updates. We therefore use $R$ executions per evaluation
case during both screening and development evaluation.

\textbf{Execution-Level Diagnostic Evidence.}
Table~\ref{tab:repeated-evidence} expands the representative ClawMark cases
discussed in Section~\ref{sec:analysis}.
Repeated executions provide useful contrasts even when aggregate task outcomes
alone appear uninformative.
When all runs succeed, differences in execution behavior can expose avoidable
overhead, as illustrated by intermittent subagent delegation.
When outcomes differ across runs, the contrast reveals trajectory-dependent
omissions that may be invisible from a single successful execution.
Conversely, failures recurring across all observed runs provide stronger
evidence of a systematic defect requiring a harness-level update.

These examples are descriptive execution evidence; because each patch may
modify multiple harness behaviors and only three executions are observed, they
should not be interpreted as isolated causal estimates of individual edits.









\begin{table}[t]
\centering
\caption{
Representative examples of how repeated executions inform harness optimization
on ClawMark. P/F denotes pass/fail of the named check.
}
\label{tab:repeated-evidence}

\begingroup
\footnotesize

\begin{tabularx}{\linewidth}{@{}
    >{\raggedright\arraybackslash}p{0.29\linewidth}
    >{\raggedright\arraybackslash}p{0.29\linewidth}
    >{\raggedright\arraybackslash}X@{}}
\toprule
\textbf{Repeated evidence}
& \textbf{Diagnostic insight}
& \textbf{Patch and repeated verification} \\
\midrule

\textbf{Costly successful path.}
A research audit scores 1.0 in all three runs, but costs
\$0.72 / \$1.58 / \$0.69.
The expensive run makes three subagent-spawn calls; the others make none.
&
Passing scores conceal different execution strategies.
The cheaper local runs provide concrete alternatives to the delegated path;
a single passing run would not reveal this contrast.
&
Prefer local execution for bounded reviews, narrow tool discovery,
and disable unavailable web tools.
After the patch, all three runs still score 1.0,
at \$0.71 / \$0.63 / \$0.81.
This is consistent with reduced overhead, not an isolated causal
estimate of delegation cost. \\
\midrule

\textbf{Medication omission.}
The medication-discrepancy check yields P/F/F:
the first execution passes, but the next two omit the required finding.
&
One successful trajectory would hide inconsistent preservation of
medication-timeline evidence as a distinct discrepancy.
&
Preserve one structured row per independent evidence issue.
Post-patch failures fall from $2/3$ to $0/3$.
A different clinical check fails once after the patch,
so targeted repair does not imply complete task correctness. \\
\midrule

\textbf{Customer-risk omission.}
The customer-champion-loss check yields P/P/F:
the missing finding becomes visible only in the third execution.
&
Combining findings into a generic risk note can lose an independent
customer-champion issue, even when other executions retain it.
&
Keep independent findings as separate evidence rows rather than
merging them into one summary.
Post-patch failures fall from $1/3$ to $0/3$. \\
\midrule

\textbf{Persistent claim-handling failure.}
The taxi-claim amount check yields F/F/F,
revealing a recurring problem across all observed executions.
&
The diagnosis identifies eligible missing-receipt claims being
omitted or rejected instead of retained for policy-permitted approval.
&
Add a policy-conditional path that retains eligible pending claims
with supporting evidence and requests approval.
Post-patch failures fall from $3/3$ to $2/3$. \\
\bottomrule
\end{tabularx}

\par\smallskip

\endgroup
\end{table}

\begin{table}[!h]
\centering
\caption{
Test results for the single-run and three-run \textsc{PS-Pareto} ablation on
ClawMark. Quality is reported in percentage (\%), and cost in USD per task.
Values are mean $\pm$ sample standard deviation across three test repeats.
Superscripts \(Q\) and \(C\) denote the quality- and cost-extreme harnesses
selected on the development set, respectively. \textbf{Bold} indicates the
best mean result in each column.
}
\label{tab:repeat-ablation}

\setlength{\tabcolsep}{4pt}
\resizebox{\textwidth}{!}{%
\begin{tabular}{
@{}l
cc
cc
cc
>{\columncolor{gray!10}}c
>{\columncolor{gray!10}}c
@{}
}
\toprule
\multirow{2}{*}{Setting}
& \multicolumn{2}{c}{E-commerce (8)}
& \multicolumn{2}{c}{Journalist (8)}
& \multicolumn{2}{c}{Legal Assistant (6)}
& \multicolumn{2}{>{\columncolor{gray!10}}c}{\textbf{Overall (22)}} \\
\cmidrule(lr){2-3}
\cmidrule(lr){4-5}
\cmidrule(lr){6-7}
\cmidrule(lr){8-9}
& Qual. $\uparrow$ & Cost $\downarrow$
& Qual. $\uparrow$ & Cost $\downarrow$
& Qual. $\uparrow$ & Cost $\downarrow$
& \textbf{Qual. $\uparrow$} & \textbf{Cost $\downarrow$} \\
\midrule

Single-run$^{C}$
& 72.46 $\pm$ 9.90
& 0.89 $\pm$ 0.02
& 82.34 $\pm$ 3.13
& \textbf{1.05} $\pm$ 0.09
& 75.74 $\pm$ 5.15
& 0.88 $\pm$ 0.09
& 76.95 $\pm$ 5.50
& 0.95 $\pm$ 0.06 \\

Single-run$^{Q}$
& 65.87 $\pm$ 4.11
& 0.83 $\pm$ 0.06
& 85.70 $\pm$ 1.57
& 1.19 $\pm$ 0.10
& 75.51 $\pm$ 7.50
& 0.87 $\pm$ 0.09
& 75.71 $\pm$ 2.10
& 0.97 $\pm$ 0.05 \\

\addlinespace[2pt]

Three-run$^{C}$
& 72.99 $\pm$ 4.29
& \textbf{0.81} $\pm$ 0.01
& \textbf{89.88} $\pm$ 3.18
& 1.12 $\pm$ 0.16
& 74.72 $\pm$ 2.95
& \textbf{0.83} $\pm$ 0.09
& 79.60 $\pm$ 2.43
& \textbf{0.92} $\pm$ 0.08 \\

Three-run$^{Q}$
& \textbf{77.77} $\pm$ 7.51
& 0.87 $\pm$ 0.07
& 87.81 $\pm$ 3.55
& 1.28 $\pm$ 0.03
& \textbf{77.51} $\pm$ 5.88
& \textbf{0.83} $\pm$ 0.08
& \textbf{81.35} $\pm$ 3.11
& 1.01 $\pm$ 0.02 \\

\bottomrule
\end{tabular}
}
\end{table}
\textbf{Single-Run Ablation Protocol.}
To test whether the additional execution evidence affects the optimization
outcome, we compare the default three-run setting ($R=3$) with a single-run
variant ($R=1$) of \textsc{PS-Pareto}. With $R=1$, the bootstrap weights are deterministic, so promotion
reduces to checking the local quality--cost condition on the
single-run observations.
The initial harness, task splits, optimizer, deployment setting, and remaining
optimization hyperparameters are held fixed; only the number of executions
used to construct and evaluate each update is changed.
Consequently, the ablation measures the end-to-end effect of optimizing with
repeated evidence, including its influence on diagnosis, patch verification,
and subsequent promotion decisions.
As reported in Table~\ref{tab:repeat-ablation}, the three-run search reaches
stronger held-out quality--cost operating points at both selected Pareto
extremes. The comparison does not isolate repeated evidence from its additional
execution cost: using $R=3$ necessarily allocates more task executions than
$R=1$. We therefore interpret the ablation as evaluating the optimization
protocol enabled by repeated execution, rather than as a compute-matched
estimate of the marginal value of each additional rollout.
\section{Additional Analysis for RQ3}
\label{app:RQ3}

\textbf{Mean-based screening.}
We compare Bayesian screening with a mean-based gate using the
same screening evidence and local allowances.
Following Section~\ref{sec:method}, let $h_t^{-}=h_t$ and $h_t^{+}=h'_t$
denote the parent and candidate, respectively.
Using the empirical estimates defined in Section~\ref{sec:formulation}, we write
\[
\widehat{\Delta Q}_t
=
\widehat{Q}_{D_t^{\mathrm{scr}}}(h_t^{+})
-
\widehat{Q}_{D_t^{\mathrm{scr}}}(h_t^{-}),
\qquad
\widehat{\Delta C}_t
=
\frac{
\widehat{C}_{D_t^{\mathrm{scr}}}(h_t^{+})
-
\widehat{C}_{D_t^{\mathrm{scr}}}(h_t^{-})
}{
\widehat{C}_{D_t^{\mathrm{scr}}}(h_t^{-})
}.
\]
Thus, $\widehat{\Delta Q}_t$ measures the absolute quality
difference, while $\widehat{\Delta C}_t$ measures the relative
cost change. The mean-based gate promotes $h'_t$ when
\[
\bigl[
\widehat{\Delta Q}_t > 0
\land
\widehat{\Delta C}_t \leq \epsilon_C
\bigr]
\;\lor\;
\bigl[
\widehat{\Delta C}_t < 0
\land
\widehat{\Delta Q}_t \geq -\epsilon_Q
\bigr].
\]

Both gates use $\epsilon_Q=0.10$ and $\epsilon_C=+\infty$.

Table~\ref{tab:replay} reports repeated observations for archived
candidate I-12 and its parent. The mean-based gate would promote
I-12, whereas its bootstrap promotion support of 0.55 falls below $\gamma=0.7$. The end-to-end gate
ablation in Section~\ref{sec:analysis} separately evaluates
how the choice of screening rule affects the resulting
optimization outcomes.






\begin{table}[!h]
\caption{
Repeated quality--cost observations for I-12 and its parent.
Brackets contain the three training executions for each task.
Differences are computed from the corresponding mean observations as
candidate minus parent; quality differences are reported in percentage
points (pp).
}
\label{tab:replay}
\centering
\small
\resizebox{\linewidth}{!}{%
\begin{tabular}{lcccccc}
\toprule
\multirow{2}{*}{Task}
& \multicolumn{2}{c}{Parent}
& \multicolumn{2}{c}{Candidate}
& \multicolumn{2}{c}{Difference} \\
\cmidrule(lr){2-3}
\cmidrule(lr){4-5}
\cmidrule(lr){6-7}
& Quality (\%) & Cost (\$)
& Quality (\%) & Cost (\$)
& $\Delta Q$ (pp) & $\Delta C$ (\$) \\
\midrule

Content operation
& $[17,\ 100,\ 100]$
& $[2.89,\ 1.26,\ 1.11]$
& $[100,\ 100,\ 100]$
& $[1.44,\ 2.08,\ 1.94]$
& $+27.67$
& $+0.067$ \\

Research assistant 4
& $[100,\ 100,\ 100]$
& $[0.65,\ 1.34,\ 1.69]$
& $[13,\ 67,\ 87]$
& $[0.76,\ 0.76,\ 0.91]$
& $-44.33$
& $-0.417$ \\

Research assistant 9
& $[100,\ 100,\ 100]$
& $[0.95,\ 0.53,\ 0.65]$
& $[100,\ 100,\ 100]$
& $[1.03,\ 0.63,\ 0.72]$
& $0.00$
& $+0.083$ \\

\midrule
\textbf{Overall mean}
& 90.74
& 1.230
& 85.19
& 1.142
& $-5.55$
& $-0.088$ \\

\bottomrule
\end{tabular}
}
\end{table}

\begin{figure}[h!]
    \centering
    \includegraphics[width=0.5\linewidth]{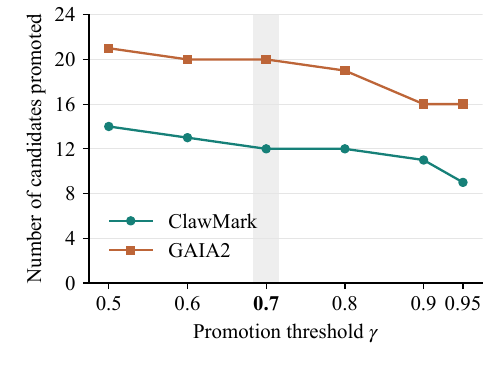}
    \caption{
    \textbf{Retrospective sensitivity to the promotion threshold $\gamma$.} Each point shows the number of archived proposals that would
    receive development evaluation under the corresponding threshold.
    The shaded region marks the setting used in our experiments.
    }
    \label{fig:bayesian_threshold}
\end{figure}
\textbf{Sensitivity to the promotion threshold.}
We replay archived proposals from both benchmarks under different
promotion thresholds \(\gamma\) (Figure~\ref{fig:bayesian_threshold}).
The setting used throughout our experiments, \(\gamma=0.7\), lies in a
locally stable region on both benchmarks, suggesting that it provides a
reasonable default. The threshold
controls how much statistical support is required before allocating broader
evaluation and does not alter the underlying quality--cost preference. For a
substantially different deployment domain, however, a small grid search over
\(\gamma\) using training or development evidence may be beneficial when
additional optimization budget is available.

\section{Optimization Stability}
\label{app:stability}

Because each full optimization run requires substantial computation,
we conduct one additional \textsc{PS-Pareto} optimization run on GAIA2
under the same experimental configuration.
Table~\ref{tab:gaia2_moo_stability} compares the original run reported
in the main text (Run~1) with this additional run (Run~2).
For each run, we select the quality- and cost-extreme harnesses on
the development set and evaluate with three repeated executions.
Both runs improve quality and reduce cost relative to the default agent across all three held-out universes.
Overall quality ranges from 61.44\% to 62.89\%, with costs of
\$0.44--\$0.47 per task, compared with 56.22\% and \$0.78 for the
default agent. Both runs also outperform all evaluated baselines
in overall mean quality and cost at both development-selected
endpoints. These results provide evidence that the
quality--cost gains persist across optimization runs.

\begin{table}[!h]
\centering
\caption{
GAIA2 test results across two \textsc{PS-Pareto} optimization runs.
Quality is reported in \%, and cost in USD per task.
Values are mean $\pm$ sample standard deviation across three test
repeats of each fixed harness. Superscripts $Q$ and $C$ denote
development-selected quality- and cost-extreme harnesses.
The shared default-agent evaluation is shown once.
}
\label{tab:gaia2_moo_stability}

\small
\setlength{\tabcolsep}{4pt}
\renewcommand{\arraystretch}{1.08}
\begin{tabular}{@{}llc>{\columncolor{gray!10}}cccc@{}}
\toprule
\multirow{2}{*}{Run}
& \multirow{2}{*}{Setting}
& \multirow{2}{*}{Metric}
& \textbf{Overall}
& \multicolumn{3}{c}{\textbf{GAIA2 by Universe}} \\
\cmidrule(lr){5-7}
& & & \textbf{GAIA2 (300)}
& U21 (107) & U22 (112) & U27 (81) \\
\midrule

\multirow{2}{*}{--}
& \multirow{2}{*}{Default Agent}
& Qual. $\uparrow$
& 56.22 $\pm$ 1.07
& 58.26 $\pm$ 1.08
& 53.57 $\pm$ 1.55
& 57.20 $\pm$ 2.57 \\
& & Cost $\downarrow$
& 0.78 $\pm$ 0.08
& 0.82 $\pm$ 0.18
& 0.71 $\pm$ 0.03
& 0.82 $\pm$ 0.04 \\
\midrule

\multirow{4}{*}{Run 1}
& \multirow{2}{*}{\textsc{PS-Pareto}$^{C}$}
& Qual. $\uparrow$
& 62.11 $\pm$ 2.46
& 63.24 $\pm$ 4.80
& 59.23 $\pm$ 1.03
& 64.61 $\pm$ 3.77 \\
& & Cost $\downarrow$
& 0.47 $\pm$ 0.00
& 0.44 $\pm$ 0.01
& 0.49 $\pm$ 0.02
& 0.47 $\pm$ 0.00 \\
& \multirow{2}{*}{\textsc{PS-Pareto}$^{Q}$}
& Qual. $\uparrow$
& 62.89 $\pm$ 2.14
& 61.99 $\pm$ 2.35
& 60.71 $\pm$ 2.36
& 67.08 $\pm$ 1.89 \\
& & Cost $\downarrow$
& 0.45 $\pm$ 0.02
& 0.39 $\pm$ 0.01
& 0.49 $\pm$ 0.03
& 0.46 $\pm$ 0.02 \\
\midrule

\multirow{4}{*}{Run 2}
& \multirow{2}{*}{\textsc{PS-Pareto}$^{C}$}
& Qual. $\uparrow$
& 61.44 $\pm$ 1.84
& 60.12 $\pm$ 1.08
& 61.61 $\pm$ 6.25
& 62.96 $\pm$ 4.28 \\
& & Cost $\downarrow$
& 0.46 $\pm$ 0.02
& 0.38 $\pm$ 0.01
& 0.47 $\pm$ 0.03
& 0.53 $\pm$ 0.02 \\
& \multirow{2}{*}{\textsc{PS-Pareto}$^{Q}$}
& Qual. $\uparrow$
& 62.22 $\pm$ 1.84
& 61.06 $\pm$ 3.28
& 61.61 $\pm$ 3.22
& 64.61 $\pm$ 1.43 \\
& & Cost $\downarrow$
& 0.44 $\pm$ 0.02
& 0.40 $\pm$ 0.03
& 0.45 $\pm$ 0.04
& 0.49 $\pm$ 0.01 \\
\bottomrule
\end{tabular}%
\end{table}

\section{Optimization Efficiency}
\label{app:optimization_efficiency}

\paragraph{Optimizer-Side Usage.}
Table~\ref{tab:gaia2_optimizer_cost} reports optimizer-side resource
usage for the GAIA2 experiments. Patch counts are totals for each
optimization run, while runtime, estimated monetary cost, LLM calls,
and token usage are averaged per generated patch.
Fresh-input, cache-read, and output tokens are reported separately.
\textsc{PS-Pareto} incurs higher estimated optimizer
cost per patch than \textsc{AS-Quality} and \textsc{MH-Pareto}
(\$9.62 versus \$6.01 and \$5.42, respectively).
This additional overhead accompanies stronger held-out results:
both development-selected \textsc{PS-Pareto} endpoints achieve
higher mean quality at lower execution cost than all evaluated
baselines. In particular, \textsc{PS-Pareto}$^{Q}$ achieves
62.89\% quality at \$0.45 per task, compared with 57.33\% at
\$0.61 for \textsc{AS-Quality} and 54.33\% at \$0.83 for
\textsc{MH-Pareto}$^{Q}$.
Similarly, \textsc{PS-MinCost} uses more optimizer cost per patch
than \textsc{MH-MinCost} (\$8.02 versus \$3.58), but achieves
higher test quality (56.11\% versus 39.78\%) at lower execution
cost (\$0.44 versus \$0.62 per task).
\begin{table}[!h]
\centering
\caption{
\textbf{Optimizer-side resource usage on GAIA2.}
Generated patches are counted per optimization run.
All remaining quantities are averaged per generated patch.
}
\label{tab:gaia2_optimizer_cost}

\small
\setlength{\tabcolsep}{4pt}
\renewcommand{\arraystretch}{1.10}
\resizebox{\textwidth}{!}{%
\begin{tabular}{@{}lrrrrrrr@{}}
\toprule
Method
& Generated
& Time (s)
& Est.\ cost (\$)
& LLM calls
& Output tokens
& Fresh-input tokens
& Cache-read tokens \\
& patches
& /patch
& /patch
& /patch
& /patch
& /patch
& /patch \\
\midrule
\textsc{AS-Quality}
& 25 & 604 & 6.01 & 63.4
& 47,331 & 391,870 & 5,260,513 \\

\textsc{MH-MinCost}
& 15 & 610 & 3.58 & 67.6
& 36,055 & 225,106 & 2,749,406 \\

\textsc{MH-Pareto}
& 15 & 619 & 5.42 & 103.3
& 50,100 & 328,311 & 4,544,512 \\

\textsc{PS-MinCost}
& 25 & 967 & 8.02 & 87.4
& 63,577 & 489,543 & 7,339,602 \\

\textsc{PS-Pareto}
& 25 & 982 & 9.62 & 86.6
& 68,526 & 616,012 & 8,968,786 \\
\bottomrule
\end{tabular}%
}
\end{table}

\section{Optimization Trajectories of \framework}
\label{app:trajectories}

\begin{figure}[!h]
    \centering
    \includegraphics[width=0.9\linewidth]{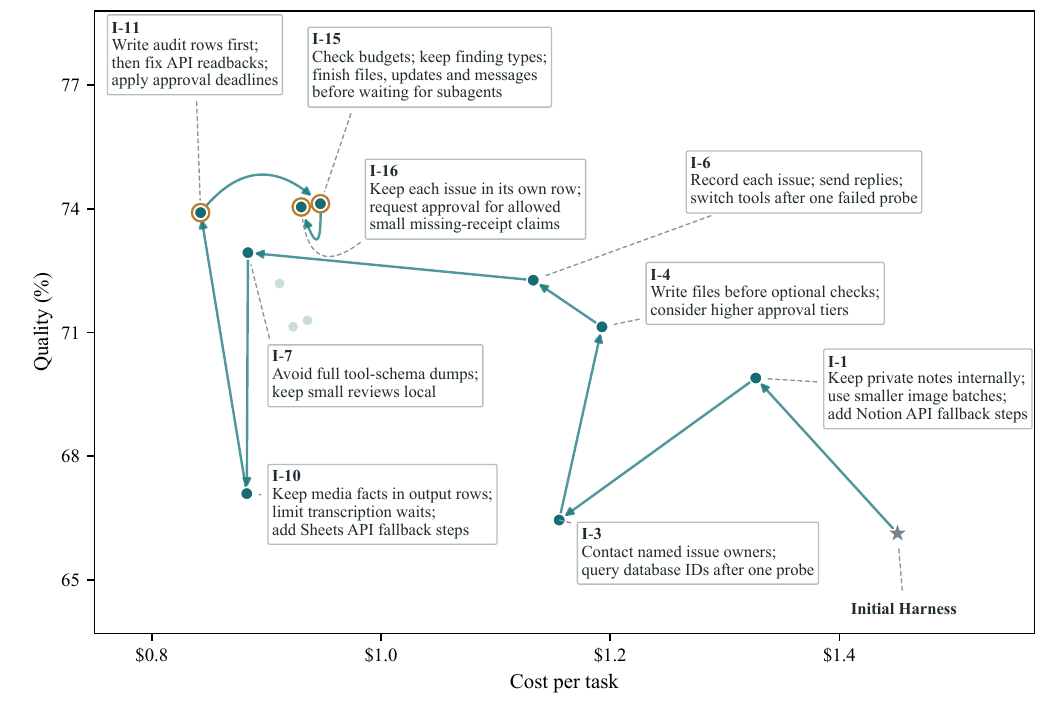}
    \caption{\textbf{Annotated development-set optimization trajectory
    of ParetoSaddler on ClawMark.}
    Solid arrows trace harness evolution; outlined markers
    identify the final development Pareto set.
    Labels indicate proposal iterations, and annotations
    summarize selected harness modifications.
    }
    \label{fig:clawmark-trajectory}
\end{figure}

\begin{figure}[!h]
    \centering
    \includegraphics[width=0.9\linewidth]{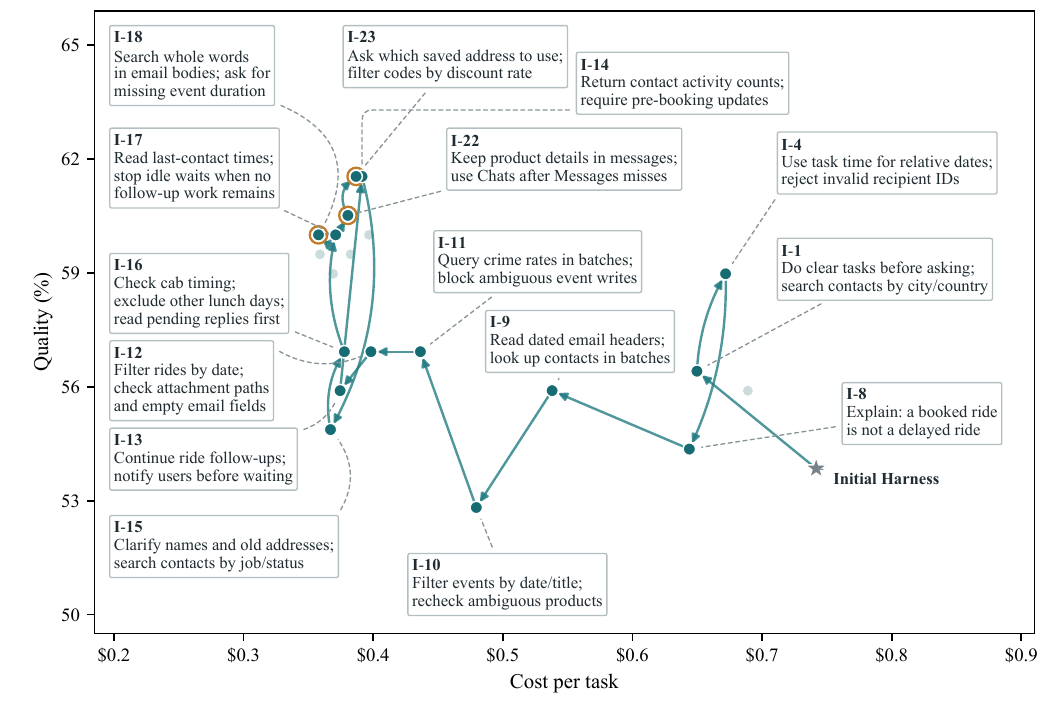}
    \caption{\textbf{Annotated development-set optimization trajectory
    of ParetoSaddler on GAIA2.}
    Solid arrows trace harness evolution; outlined markers
    identify the final development Pareto set.
    Labels indicate proposal iterations, and annotations
    summarize selected harness modifications.
    }
    \label{fig:gaia2-trajectory}
\end{figure}

Figures~\ref{fig:clawmark-trajectory}
and~\ref{fig:gaia2-trajectory} visualize the development
quality--cost trajectories of \textsc{PS-Pareto} on ClawMark
and GAIA2, respectively. Iteration labels identify candidates
within each optimization run, while annotations summarize
selected harness modifications. All plotted outcomes are
development estimates; final held-out results are reported
separately in the main text.

\paragraph{ClawMark.}
The trajectory includes updates to evidence preservation,
tool access, and task completion.
Early modifications address omitted information and unsuccessful
tool interactions. I-7 introduces bounded probing and local-first
execution, improving development quality while reducing cost
relative to its parent. Later candidates add guidance for audit
artifacts, budget exceptions, and independent evidence rows.
The final development Pareto archive contains I-11, I-15, and
I-16, retaining multiple quality--cost alternatives for deployment
selection.

\paragraph{GAIA2.}
The annotated updates address information retrieval, ambiguity
resolution, and execution control. I-9 reduces retrieval overhead
through compact email reads and batched contact lookup, while
I-17 improves access to email history and limits unnecessary
waiting after task completion. As detailed in
Appendix~\ref{app:rq1}, both candidates retain perfect screening
quality while reducing cost and subsequently improve both
development objectives. These cases illustrate how cost
reductions can provide a useful promotion signal when local
quality is saturated.

The trajectories show that successive updates need not improve
both objectives simultaneously, and a candidate can become
dominated as later alternatives emerge. The annotations describe
implemented changes rather than isolated causal explanations
of performance: patches may modify several behaviors, and
execution outcomes are stochastic. The controlled gate ablation
in Section~5.3 provides complementary evidence about the effect
of the promotion rule.
\section{Complete ClawMark Optimization Run}
\label{app:clawmark-search-trace}

We document all 17 proposals from one ClawMark \textsc{PS-Pareto}
run of \framework.
The development split contains 20 tasks, yielding 60 rollouts per
full-development evaluation.
Table~\ref{tab:clawmark-search-trace} summarizes the optimizer's trajectory-based
diagnoses and implemented changes. These diagnoses are interpretations of
execution feedback rather than independently established causal explanations.
Of the 17 proposals, 12 received full-development evaluation and five were
rejected. The final archive contains I-11, I-15, and I-16, with
I-16 selected as the final current candidate.

\begingroup
\setlength{\tabcolsep}{4pt}
\begin{longtable}{@{}
>{\raggedright\arraybackslash}p{0.05\textwidth}
>{\raggedright\arraybackslash}p{0.36\textwidth}
>{\raggedright\arraybackslash}p{0.37\textwidth}
>{\raggedright\arraybackslash}p{0.15\textwidth}@{}}
\caption{Complete ClawMark multi-objective search trace. Rejected candidates did not receive full-development evaluation during optimization. Dominated and Pareto refer to final development status within this search, not status when first evaluated. Prompt and skill guidance describes intended behavior, not an enforced execution guarantee.}
\label{tab:clawmark-search-trace}\\
\toprule
Iter & Diagnosed issue & Implemented change & Final status \\
\midrule
\endfirsthead
\multicolumn{4}{c}{Table \thetable\ continued}\\
\toprule
Iter & Diagnosed issue & Implemented change & Final status \\
\midrule
\endhead
\midrule
\multicolumn{4}{r}{Continued on next page}\\
\endfoot
\bottomrule
\endlastfoot
1 & Private allergy notes and same-name ambiguity were overlooked; schema probing, unavailable Python commands, and large image batches consumed budget. & Added internal-evidence and Python compatibility guidance, Notion raw-API fallback steps, and smaller-batch advice in the image-tool description. & Dominated \\
\midrule
2 & Live trackers were not updated; budgets omitted actual spending; release conclusions omitted observed log gaps; schema and package probing continued after sufficient evidence. & Added guidance for live updates, complete budget arithmetic, release-evidence preservation, and fallbacks without extra dependencies. & Rejected \\
\midrule
3 & Required mentor contact was omitted; repeated Notion schema probes continued despite known identifiers and usable raw APIs. & Added responsible-party contact guidance and instructions to use known identifiers or raw APIs after one schema probe. & Dominated \\
\midrule
4 & Optional presentation and media exploration delayed required artifacts; a higher compensation tier was conflated with a compliance hold. & Added artifact-first guidance and instructions to report provisional tier eligibility separately while preserving compliance holds. & Dominated \\
\midrule
5 & Missing memory tools and helpers were repeatedly probed; a disputed appeal was prematurely marked resolved. & Added bounded-fallback and unresolved-appeal guidance. & Rejected \\
\midrule
6 & Contradictions and independent expense flags were omitted; requested financial messages were only drafted; missing helpers triggered retries. & Added row-level discrepancy and communication guidance, and bounded helper retries. & Dominated \\
\midrule
7 & Unavailable web tools, full-schema reads, helper probes, and delegation of bounded local work added overhead despite high training scores. & Disabled unavailable web tools in configuration; revised Notion and tool-access skills to avoid full-schema reads; added local-first and bounded-fallback guidance. & Dominated \\
\midrule
8 & Optional web exploration continued despite sufficient local evidence; one rollout exhausted its length budget without delivery; scheduling, risk calibration, and transcription polling were inefficient. & Added local-evidence, date, and risk-calibration guidance; revised Whisper skill instructions to start with a small model and limit polling. & Dominated \\
\midrule
9 & Time-zone and commute constraints were not jointly checked; optional meeting setup delayed output; invitations violated an explicit eligibility threshold. & Added scheduling and hard-threshold guidance, artifact-before-setup instructions, and local handling of bounded work. & Dominated \\
\midrule
10 & Visual facts and unassigned actions were lost; transcription and process-control loops displaced deliverables; live spreadsheet writes were difficult to verify. & Added media-evidence and live-readback guidance, a Sheets REST fallback in tool-use instructions, and targeted transcription with bounded polling. & Dominated \\
\midrule
11 & Explicit attendance evidence remained pending; backend exploration delayed required audit and reconciliation tables. & Added local-audit-first guidance, followed by live writes and readback, and policy-timestamp-based status decisions. & Pareto \\
\midrule
12 & Bounded record review was delegated, and execution yielded before required local artifacts were written. & Reworked subagent-orchestration guidance to prioritize local artifacts, delegate only separable work, and yield only when necessary. & Rejected \\
\midrule
13 & HR outlier counts and current recruiting stages were inconsistent; an explicitly required action remained only in notes. & Added outlier-counting, live/local consistency, and independent action-row guidance for explicit instructions. & Rejected \\
\midrule
14 & Optional backend and public queries continued after sufficient evidence; date discrepancies and budget overruns were not reported as exceptions. & Added evidence-sufficiency stopping guidance and explicit cross-document date, status, and budget discrepancy reporting. & Rejected \\
\midrule
15 & Budget flags were omitted; waiting for subagents blocked email delivery; generic summaries replaced schema-specific findings. & Added budget-exception and schema-specific evidence guidance; required feasible artifacts, live writes, and communications before yielding. & Pareto \\
\midrule
16 & An aspirin timeline and a stakeholder's departure were not preserved as independent evidence; a low-value taxi claim without a receipt was rejected despite an approval exception. & Added schema-aligned evidence-row guidance and instructions to request approval for eligible missing-receipt claims rather than silently reject them. & Pareto \\
\midrule
17 & Recommendations were confused with purchase authorization; primary blockers were omitted; unavailable helpers and invalid Notion URLs disrupted execution. & Added recommendation-versus-authorization and blocker-preservation guidance, with immediate fallback after invalid Notion URLs. & Dominated \\
\end{longtable}
\endgroup

\paragraph{Final Harness Changes.}
We summarize the cumulative changes from the initial harness
to I-16, the latest candidate in the final Pareto
archive. Table~\ref{tab:clawmark-i16-harness-changes} covers
six skill and source file changes, grouped by directory.
All paths are relative to the OpenClaw repository root.

\begingroup
\small
\setlength{\tabcolsep}{5pt}
\setlength{\LTleft}{0pt}
\setlength{\LTright}{0pt}
\renewcommand{\arraystretch}{1.12}
\begin{longtable}{@{}
    >{\raggedright\arraybackslash}p{0.30\linewidth}
    >{\raggedright\arraybackslash}p{\dimexpr0.70\linewidth-10pt\relax}
@{}}
\caption{Cumulative harness-file changes in ClawMark I-16 relative
to the initial harness. File paths are relative to the directory
shown in each group heading.}
\label{tab:clawmark-i16-harness-changes}\\
\toprule
\textbf{File} & \textbf{Summary of changes} \\
\midrule
\endfirsthead
\multicolumn{2}{@{}l}{\small\tablename~\thetable\ (continued)}\\
\toprule
\textbf{File} & \textbf{Summary of changes} \\
\midrule
\endhead
\midrule
\multicolumn{2}{r@{}}{\footnotesize Continued on next page}\\
\endfoot
\bottomrule
\endlastfoot

\rowcolor{gray!10}

\multicolumn{2}{@{}l@{}}{\strut\textbf{Tool-use skills}\quad{\footnotesize\nolinkurl{skills/}}}\\*
\addlinespace[3pt]
\nolinkurl{gog/SKILL.md} & Adds direct Google Sheets REST API fallback when helper tools or packages are unavailable; requires reading back updated ranges to verify live writes. \\
\addlinespace[5pt]
\nolinkurl{mcporter/SKILL.md} & Discourages full schema dumps; recommends compact tool listings or inspection of one specific function. \\
\addlinespace[5pt]
\nolinkurl{notion/SKILL.md} & Adds known mcporter calls and raw-API fallback for task-provided database IDs; requires querying records before concluding and retaining relevant private evidence in internal outputs without unauthorized external disclosure. \\
\addlinespace[5pt]
\nolinkurl{openai-whisper/SKILL.md} & Recommends a small model for short English audio, targeted audio windows for long recordings, one active transcription strategy, and bounded polling before abandoning stalled jobs. \\
\addlinespace[7pt]
\rowcolor{gray!10}
\multicolumn{2}{@{}l@{}}{\strut\textbf{Agent and tool prompts}\quad{\footnotesize\nolinkurl{src/agents/}}}\\*
\addlinespace[3pt]
\nolinkurl{system-prompt.ts} & Prioritizes required artifacts, live updates, read-back verification, and communications before optional exploration or delegation waits. Bounds helper/schema probing and media processing; favors local completion of bounded reviews. Preserves schema-specific evidence and separate independent issues, including medication timelines and customer/champion risks. Clarifies budget checks, policy-based statuses, approval tiers, and pending-approval handling for eligible missing-receipt claims. \\
\addlinespace[5pt]
\nolinkurl{tools/image-tool.ts} & Revises image-tool descriptions to recommend small, focused batches and smaller retries after weak results or timeouts; leaves image-processing implementation unchanged. \\
\end{longtable}
\endgroup

\paragraph{Complementary Case: \textsc{PS-MinCost}.}
We additionally summarize the selected harness from the ClawMark
\textsc{PS-MinCost} run using the local Pareto-aware gate.
Its selected candidate, also indexed I-16, belongs to a separate
optimization run from the \textsc{PS-Pareto} candidate above.
Table~\ref{tab:clawmark-mincost-i16-changes} summarizes its cumulative
changes relative to its initial harness across 15 skill and source
files.

\begingroup
\small
\setlength{\tabcolsep}{5pt}
\setlength{\LTleft}{0pt}
\setlength{\LTright}{0pt}
\renewcommand{\arraystretch}{1.12}
\begin{longtable}{@{}
    >{\raggedright\arraybackslash}p{0.30\linewidth}
    >{\raggedright\arraybackslash}p{\dimexpr0.70\linewidth-10pt\relax}
@{}}
\caption{Cumulative harness-file changes in ClawMark \textsc{PS-MinCost} I-16 under Pareto local promotion. File paths are relative to the directory shown in each group heading.}
\label{tab:clawmark-mincost-i16-changes}\\
\toprule
\textbf{File} & \textbf{Summary of changes} \\
\midrule
\endfirsthead
\multicolumn{2}{@{}l}{\small\tablename~\thetable\ (continued)}\\
\toprule
\textbf{File} & \textbf{Summary of changes} \\
\midrule
\endhead
\midrule
\multicolumn{2}{r@{}}{\footnotesize Continued on next page}\\
\endfoot
\bottomrule
\endlastfoot
\rowcolor{gray!10}
\multicolumn{2}{@{}l@{}}{\strut\textbf{Tool-use skills}\quad{\footnotesize\nolinkurl{skills/}}}\\*
\addlinespace[3pt]
\nolinkurl{mcporter/SKILL.md} & Discourages repeated full-schema inspection; recommends focused calls and alternative Notion endpoints or raw REST after invalid-URL failures. \\
\addlinespace[4pt]
\nolinkurl{notion/SKILL.md} & Adds sandbox-specific REST fallback guidance and directs batched operations toward supplied resource IDs rather than repeated searches or ID-format retries. \\
\addlinespace[4pt]
\nolinkurl{openai-whisper/SKILL.md} & Prioritizes existing transcripts and small transcription models; limits transcription to decision-critical audio and recommends one long wait instead of repeated polling or duplicate jobs. \\
\addlinespace[4pt]
\nolinkurl{video-frames/SKILL.md} & Recommends an initial contact sheet of four to six frames per video, followed by targeted inspection when needed; preserves visual risks, blockers, ownerless tasks, and delays as structured findings. \\
\addlinespace[7pt]
\rowcolor{gray!10}
\multicolumn{2}{@{}l@{}}{\strut\textbf{Agent tools and execution}\quad{\footnotesize\nolinkurl{src/agents/}}}\\*
\addlinespace[3pt]
\nolinkurl{agent-tools.read.ts} & Filters detected raw binary text and removes inline image blocks exceeding 24\,KiB of base64 data from read results; directs the agent to appropriate extraction or image tools. This changes evidence delivery rather than image resolution. \\
\addlinespace[4pt]
\nolinkurl{sessions/tools/read.ts} & Revises read-tool guidance to favor text files and route documents, images, and other media through dedicated extraction paths. \\
\addlinespace[4pt]
\nolinkurl{bash-tools.exec-output.ts} & Compacts recognized MCP schema dumps of at least 20,000 characters into at most 80 function signatures; appends targeted recovery hints for Notion URL errors and missing Python packages. \\
\addlinespace[4pt]
\nolinkurl{bash-tools.exec.ts} & Detects local Whisper commands and enforces at least a 45-second initial wait before background continuation; adds guidance for decision-critical transcription waits. \\
\addlinespace[4pt]
\nolinkurl{bash-tools.process.ts} & Applies shared output compaction to process polling and logs; allows Whisper polls up to 120 seconds while retaining the 30-second cap for other commands; discourages repeated empty polls and duplicate transcription. \\
\addlinespace[4pt]
\nolinkurl{bash-tools.whisper.ts} & Adds a shared detector for Whisper CLI commands, module invocations, and inline Python scripts. \\
\addlinespace[4pt]
\nolinkurl{bash-tools.schemas.ts} & Documents the distinct polling limits for ordinary processes and local Whisper jobs. \\
\addlinespace[4pt]
\nolinkurl{bash-tools.descriptions.ts} & Encourages bounded command output, fewer empty polls, and targeted long waits for necessary transcription. \\
\addlinespace[4pt]
\nolinkurl{bash-tools.shared.ts} & Defines a conditional Python compatibility shim and injects it into Docker shell commands, mapping bare \texttt{python} to \texttt{python3} only when needed. \\
\addlinespace[4pt]
\nolinkurl{bash-tools.exec-runtime.ts} & Integrates the same compatibility shim into POSIX host execution while preserving configured PATH handling. \\
\addlinespace[4pt]
\nolinkurl{system-prompt.ts} & Keeps bounded evidence-to-writeback work local; requires explicit files, backend updates, and communications. Encourages batched reads, transcript reuse, broad initial visual inspection, in-place tracker updates, correct workspace paths, and bounded dependency fallback. \\
\end{longtable}
\endgroup
\section{Complete GAIA2 Optimization Run}
\label{app:gaia2-search-trace}

We document all 25 proposals from one GAIA2 \textsc{PS-Pareto} optimization
run of \framework.
The development split contains 65 tasks, yielding 195 rollouts per
full-development evaluation.
Table~\ref{tab:gaia2-search-trace} summarizes the optimizer's trajectory-based
diagnoses and implemented changes. 

\begingroup
\setlength{\tabcolsep}{4pt}
\begin{longtable}{@{}
>{\raggedright\arraybackslash}p{0.05\textwidth}
>{\raggedright\arraybackslash}p{0.36\textwidth}
>{\raggedright\arraybackslash}p{0.37\textwidth}
>{\raggedright\arraybackslash}p{0.15\textwidth}@{}}
\caption{Complete GAIA2 search trace. Rejected candidates did not receive full-development evaluation during optimization. Dominated and Pareto refer to final development status within this search, not status when first evaluated.}
\label{tab:gaia2-search-trace}\\
\toprule
Iter & Diagnosed issue & Implemented change & Final status \\
\midrule
\endfirsthead
\multicolumn{4}{c}{Table \thetable\ continued}\\
\toprule
Iter & Diagnosed issue & Implemented change & Final status \\
\midrule
\endhead
\midrule
\multicolumn{4}{r}{Continued on next page}\\
\endfoot
\bottomrule
\endlastfoot
1 & Blocked rescheduling delayed safe deletions; announcements omitted details; contact lookup required paging. & Added partial-completion and message guidance, pagination reminders, and city/country contact search. & Dominated \\
\midrule
2 & Cancellation messages were vague; punctuation-sensitive product search caused retries. & Added self-contained message guidance and normalized fallback search after empty product results. & Dominated \\
\midrule
3 & Email replies omitted detail and signatures; bare final text caused invalid-answer retries. & Added reply-completeness guidance and a final-answer tool-use guard. & Rejected \\
\midrule
4 & Relative dates were misinterpreted; contact IDs were used as phone recipients; separate emails were batched. & Added temporal guidance, Calendar/Email contracts, and phone-recipient validation. & Dominated \\
\midrule
5 & Literal prefix matching missed required files; missing role search prompted broad contact scans. & Added role-aware contact search and multi-file cleanup guidance. & Rejected \\
\midrule
6 & Name ambiguity failed to block dependent writes; final answers incurred retries. & Added ambiguity-aware write guards, focused blocker messages, and tool-based final delivery. & Rejected \\
\midrule
7 & Unresolved event references prompted speculative alternatives and broad historical searches. & Added exact-event clarification guidance and narrow date/time search preferences. & Rejected \\
\midrule
8 & A BOOKED cab's nonzero delay estimate triggered unnecessary contingency actions. & Clarified BOOKED versus DELAYED cab semantics. & Dominated \\
\midrule
9 & Successful aggregation retained full email bodies and used serial contact lookups. & Added compact date-filtered email reads, batched contact lookup, and lightweight membership checks. & Dominated \\
\midrule
10 & Broad calendar evidence distorted attendee ranking; shopping output obscured variants. & Added scoped calendar search, variant-aware product output, and finalization guards. & Dominated \\
\midrule
11 & Daily replacements were created per cancellation; location checks were serial; internal markers polluted context. & Added cardinality guards, batched crime lookup, apartment filters, and internal-marker removal. & Dominated \\
\midrule
12 & Cab locations and meeting references remained ambiguous; email roles and tool arguments were misinterpreted. & Added ambiguity guards, date-filtered ride history, role search, and revised email/checkout contracts. & Dominated \\
\midrule
13 & Execution stopped before cab contingencies or waited before updates; questions and product searches were mis-scoped. & Added continuation, update-before-wait, and concise-question guidance, plus age-aware product search. & Dominated \\
\midrule
14 & Timeouts were treated as availability; activity counting required extensive paging. & Added availability-dependent booking guards and compact contact activity summaries. & Dominated \\
\midrule
15 & Ambiguous names and stale addresses led to unsafe actions; filtered contacts required paging. & Added recipient/address guards and status/profession contact search. & Dominated \\
\midrule
16 & Cab timing and pending replies were misinterpreted; email sign-offs were unstable. & Added timing, reply-completeness, lunch-conflict, and email-sign-off guidance. & Dominated \\
\midrule
17 & Completed shopping re-entered idle waits; email queries lacked recency and oldest-first access. & Added termination guidance, recency summaries, and oldest-first email listing. & Dominated \\
\midrule
18 & Substring matches caused extra email replies; calendar clarification omitted missing duration. & Added scoped whole-word email search and complete conflict-clarification guidance. & Pareto \\
\midrule
19 & Purchase messages omitted products; gift emails were unstable; known contacts triggered repeated lookups. & Added product-specific announcement, gift-notice, and compact messaging-availability guidance. & Dominated \\
\midrule
20 & Cancellation workflows omitted follow-up; attendance emails lacked detail; broad calendar reads increased cost. & Added follow-up and email-completeness guards and discouraged broad calendar reads. & Rejected \\
\midrule
21 & Scheduled-call replies completed writes but failed semantic checks due to underspecified wording. & Added a targeted scheduled-call reply guard and aligned tool guidance. & Dominated \\
\midrule
22 & Fallback emails omitted unanswered-message context; product descriptors were lost; fuzzy searches were redundant. & Added fallback framing, descriptor-preservation guidance, and direct Messages-to-Chats fallback. & Pareto \\
\midrule
23 & Phone recipients were treated as unavailable; apartment ambiguity was unresolved; discount inspection was serial. & Added apartment-booking guards, direct phone-number messaging, and filtered discount lookup. & Pareto \\
\midrule
24 & Combined name/city search obscured name ambiguity before destructive or dependent writes. & Added multi-field contact search and a destructive-action ambiguity guard. & Dominated \\
\midrule
25 & Location ambiguity blocked unrelated work; cab arguments mismatched expected behavior; discount discovery was repeated. & Added cab ambiguity and immediate-ride guidance, delayed-cab continuation, and threshold discount filtering. & Dominated \\
\end{longtable}
\endgroup

\paragraph{Final Harness Changes.}
We summarize the cumulative changes from the initial harness
to I-23, the latest candidate in the final Pareto
archive. Table~\ref{tab:i23-harness-changes} covers the 13 modified
harness source files, grouped by directory.
All paths are relative to \nolinkurl{are/simulation/}.

\begingroup
\small
\setlength{\tabcolsep}{5pt}
\setlength{\LTleft}{0pt}
\setlength{\LTright}{0pt}
\renewcommand{\arraystretch}{1.12}

\begin{longtable}{@{}
    >{\raggedright\arraybackslash}p{0.30\linewidth}
    >{\raggedright\arraybackslash}p{\dimexpr0.70\linewidth-10pt\relax}
@{}}

\caption{Cumulative source-file changes in I-23 relative to
the initial harness. File paths are relative to the directory
shown in each group heading.}
\label{tab:i23-harness-changes}\\

\toprule
\textbf{Source file} & \textbf{Summary of changes} \\
\midrule
\endfirsthead

\multicolumn{2}{@{}l}{%
    \small\tablename~\thetable\ (continued)}\\
\toprule
\textbf{Source file} & \textbf{Summary of changes} \\
\midrule
\endhead

\midrule
\multicolumn{2}{r@{}}{\footnotesize Continued on next page}\\
\endfoot

\bottomrule
\endlastfoot

\rowcolor{gray!10}
\multicolumn{2}{@{}l@{}}{%
    \strut\textbf{Agent control and prompts}
    \quad{\footnotesize\nolinkurl{agents/default_agent/}}}\\*
\addlinespace[3pt]

\nolinkurl{base_agent.py} &
Adds pre-tool-use guards for contact and destination
ambiguity, calendar updates and invitations, email arguments,
batched queries, product descriptors, conditional cab booking,
and notification ordering. \\
\addlinespace[5pt]

\nolinkurl{prompts/system_prompt.py} &
Directs final and clarification messages through the
user-interface tool; requires completing clear earlier
subtasks before blocked later actions; anchors relative
dates to the task or environment time. \\
\addlinespace[5pt]

\nolinkurl{termination_methods/are_simulation.py} &
Continues execution after cab-booking acknowledgements when
delay or cancellation handling remains pending; adds
termination handling for completed shopping workflows
requesting no further notifications. \\
\addlinespace[7pt]

\rowcolor{gray!10}
\multicolumn{2}{@{}l@{}}{%
    \strut\textbf{Application tools}
    \quad{\footnotesize\nolinkurl{apps/}}}\\*
\addlinespace[3pt]

\nolinkurl{email_client.py} &
Adds date filters, header-only results, oldest-first listing,
scoped whole-word search, and batched contact email activity
and recency summaries; refines email-sending instructions. \\
\addlinespace[5pt]

\nolinkurl{contacts.py} &
Supports multiple contact queries in one search and extends
matching to city, country, status, and more precise job or
role information. \\
\addlinespace[5pt]

\nolinkurl{calendar.py} &
Adds time-bounded tag lookup and event search with optional
title-only matching; refines calendar-operation instructions. \\
\addlinespace[5pt]

\nolinkurl{messaging_v2.py} &
Adds batched contact activity summaries, strengthens
phone-number validation to reject internal contact IDs,
and clarifies recipient lookup, pagination, and
message-content requirements. \\
\addlinespace[5pt]

\nolinkurl{messaging.py} &
Clarifies pagination limits and message composition,
including concise requested questions and preservation
of event details and product-variant descriptors. \\
\addlinespace[5pt]

\nolinkurl{shopping.py} &
Extends product search to variant attributes with token
normalization and age matching; revises product and item
representations; adds discount-code filtering by percentage. \\
\addlinespace[5pt]

\nolinkurl{apartment_listing.py} &
Adds unsaved-property filtering and multi-zip-code search;
instructs the agent to treat multiple matching saved properties
as ambiguous for location-dependent actions. \\
\addlinespace[5pt]

\nolinkurl{cab.py} &
Adds pagination and date filtering to ride-history lookup;
revises ride representation and guidance for quotations,
bookings, and status checks, including destination ambiguity. \\
\addlinespace[5pt]

\nolinkurl{city.py} &
Supports batched crime-rate queries for multiple zip codes. \\
\addlinespace[5pt]

\nolinkurl{system.py} &
Clarifies that required status updates precede waiting
and that completed tasks should not end with an unnecessary
idle wait. \\

\end{longtable}
\endgroup

\paragraph{Complementary Case: \textsc{PS-MinCost}.}
We additionally summarize the selected harness from the GAIA2
\textsc{PS-MinCost} run with the local Pareto-aware gate.
Its selected candidate, I-22, belongs to a separate optimization run
from the \textsc{PS-Pareto} candidate above.
Table~\ref{tab:gaia2-mincost-i22-changes} summarizes its cumulative
changes relative to its own initial harness across 12 Python source
files and one runtime hook configuration.

\begingroup
\small
\setlength{\tabcolsep}{5pt}
\setlength{\LTleft}{0pt}
\setlength{\LTright}{0pt}
\renewcommand{\arraystretch}{1.12}
\begin{longtable}{@{}
    >{\raggedright\arraybackslash}p{0.30\linewidth}
    >{\raggedright\arraybackslash}p{\dimexpr0.70\linewidth-10pt\relax}
@{}}
\caption{Cumulative harness-file changes in GAIA2 \textsc{PS-MinCost} I-22 with local Pareto promotion. File paths are relative to the directory shown in each group heading.}
\label{tab:gaia2-mincost-i22-changes}\\
\toprule
\textbf{File} & \textbf{Summary of changes} \\
\midrule
\endfirsthead
\multicolumn{2}{@{}l}{\small\tablename~\thetable\ (continued)}\\
\toprule
\textbf{File} & \textbf{Summary of changes} \\
\midrule
\endhead
\midrule
\multicolumn{2}{r@{}}{\footnotesize Continued on next page}\\
\endfoot
\bottomrule
\endlastfoot
\rowcolor{gray!10}
\multicolumn{2}{@{}l@{}}{\strut\textbf{Agent execution and prompts}}\\*
\multicolumn{2}{@{}l@{}}{\footnotesize\nolinkurl{are/simulation/agents/default_agent/}}\\*
\addlinespace[3pt]
\nolinkurl{base_agent.py} & Adds heuristic conversion of terminal plain-text responses into user-interface tool calls after prior tool activity; checks intended to exclude refusals and progress messages limit when conversion applies. \\ \addlinespace[4pt]
\nolinkurl{prompts/additional_prompts.py} & Prioritizes compact search, filtering, and counting over broad listing and repeated lookups; requires tool-based final responses and clarifications; clarifies authorization for simulated app actions. \\ \addlinespace[7pt]
\rowcolor{gray!10}
\multicolumn{2}{@{}l@{}}{\strut\textbf{Application tools}\quad{\footnotesize\nolinkurl{are/simulation/apps/}}}\\*
\addlinespace[3pt]
\nolinkurl{apartment_listing.py} & Adds multi-zip-code search, unsaved-property filtering, and sorting by square footage; refines listing and saving guidance. \\ \addlinespace[4pt]
\nolinkurl{cab.py} & Adds compact ride-history search by destination, multiple candidate destinations, origin, and status; refines time formatting and booking guidance. \\ \addlinespace[4pt]
\nolinkurl{calendar.py} & Adds compact multi-tag event lookup and attendee-based event counting with date filters; refines event-query and ambiguity-handling guidance. \\ \addlinespace[4pt]
\nolinkurl{city.py} & Adds batched crime-rate lookup and threshold-based zip-code filtering. \\ \addlinespace[4pt]
\nolinkurl{contacts.py} & Introduces compact contact representations, batched full-name lookup, age filtering, and exact residence and nationality filters; adds ambiguous-recipient guidance. \\ \addlinespace[4pt]
\nolinkurl{email_client.py} & Adds date, sender, and recipient filters, summary-only results, field-restricted whole-word search, and normalized address matching; revises email-sending and reply instructions. \\ \addlinespace[4pt]
\nolinkurl{messaging_v2.py} & Adds date-window conversation reads and participant-name-aware search; refines recipient lookup, direct reply checking, and product-availability message requirements. \\ \addlinespace[4pt]
\nolinkurl{sandbox_file_system.py} & Adds filtered file discovery by filename prefixes, substrings, or suffixes without returning full directory trees; clarifies directory creation and existence checks. \\ \addlinespace[4pt]
\nolinkurl{shopping.py} & Extends normalized product matching, adds item-specific discount-code lookup, and encourages reuse of available product details and known cart state. \\ \addlinespace[4pt]
\nolinkurl{system.py} & Instructs the agent to execute recurring actions at their scheduled intervals through the final required interval before proceeding to later checks. \\ \addlinespace[7pt]
\rowcolor{gray!10}
\multicolumn{2}{@{}l@{}}{\strut\textbf{Runtime configuration}\quad{\footnotesize Repository root}}\\*
\addlinespace[3pt]
\nolinkurl{hook.json} & Adds five pre-tool-use reminders covering recipient ambiguity, direct conversation reads, calendar monitoring, item-specific discounts, and reuse of product details. The calendar reminder references \texttt{get\_events\_added\_by\_others}, which is not implemented in the archived harness; it is not an added tool capability. \\
\end{longtable}
\endgroup

\section{Optimizer Prompts}
\label{app:prompts}

Building on AutoSaddler's multi-session optimizer, ParetoSaddler extends
the session instructions to account for both response quality and execution
cost. We present selected additions and replacements, relative to the corresponding
quality-only templates.
The boxes reproduce selected passages with line wrapping adjusted;
\texttt{[...]} marks omissions, and braces denote runtime placeholders.
Promotion and archive decisions are computed by the outer
optimization loop, rather than by the optimizer's textual assessment.

\paragraph{Candidate Preparation (Session 0).}
The evolution session uses development-set quality--cost measurements
and archive membership to prepare the next candidate. In the template
shown here, the outer loop fixes the current parent; the optimizer may
repair uncompensated regressions or transfer localized mechanisms from
eligible archive members.

\begin{promptbox}{Session 0: Candidate Preparation --- Modified Instructions}
## Goal

Prepare the already-forked current harness source tree for this iteration's multi-objective optimization patch. The objective is to grow the Pareto frontier by maximizing accuracy and minimizing cost. The outer loop has already selected the `CURRENT` base; Session 0 must not reselect or replace it.

Preparation must be **data-driven**, based on full-development objective vectors, MOO status, outcome analysis, execution traces, and focused source comparison - not attachment to accumulated patches or a single-objective ranking.

[...]

## Candidate Performance

{candidate_table}

[...]

## Preparation Strategy

The current child is always the starting point. Choose among these actions:

- **Keep the current fork unchanged** when no evidence-backed Pareto-relevant preparation is warranted.
- **Repair an inherited Pareto regression** by reverting only lines that caused quality loss or cost growth without a compensating objective gain.
- **Transfer a Pareto mechanism** when a donor has a measured, localized quality, cost, or joint improvement that complements the current child.

[...]

The resulting source must remain a descendant of the fixed current fork. Donor use adds provenance; it does not change the base parent.
\end{promptbox}

\paragraph{Diagnosis--Patch (Session 1).}
Diagnosis is extended from unsuccessful tasks to cost-inefficient executions, including tasks that already pass. The session receives measured cost and per-run evidence alongside quality scores, and is instructed to inspect repeated executions before making causal or variance claims. The objective permits quality improvements, cost improvements, and trade-offs, while explicitly guarding against savings obtained by omitting required work.

\begin{promptbox}{Session 1: Diagnosis--Patch --- Modified Instructions}
## Goal

Fix incomplete or cost-inefficient scenarios in this mini-batch by diagnosing root causes from execution traces and the agent codebase, then applying targeted code patches without regressing scenarios that already pass.

[...]

The optimization objective is to grow a useful **quality-cost Pareto frontier** across the full task distribution, not just this mini-batch. Joint improvements, quality-improving, or cost-improving are all valid. Patches should address general behavioral patterns, not hardcode scenario-specific answers.

[...]

Do not create cheap failure by removing required work or validation.

[...]

## Initial Measured Cost
{cost_listing}

### Cost breakdown - every task

Token and artifact breakdown for **each** task below. When repeated runs are enabled, each task also lists per-run score, cost, token counters, and trace paths.
{cost_breakdown}

[...]

### 2. Diagnose failing or expensive scenarios

For each failing scenario or expensive run, use the `diagnose` skill to identify the root cause. The key is to find **why** the agent made the wrong decision or became costly, not just **what** went wrong. Diagnosis requires reading both the execution trace and the agent codebase.

Use the exact per-run artifact paths in the cost breakdown to inspect harness output. Start with `run/output.jsonl` for score, status, rationale, and exception state; then read the matching lite trace for agent behavior and input/cache/output token usage. Escalate to the matching full trace for
environment, app-state, event, or judge evidence. For `num_runs > 1`, inspect every retained run before making causal or variance claims; for `num_runs = 1`, the single row and trace are the complete run-level evidence. Use only these benchmark artifacts for execution and cost attribution.
\end{promptbox}

\paragraph{Reflection (Session 2).}
Reflection separately records measured objective movement, evidence that the intended mechanism was exercised, and attribution to the patch or execution variability. It asks whether additional work supports quality and whether cost reductions preserve required behavior.

\begin{promptbox}{Session 2: Reflection --- Modified Instructions}
## Objective history (accuracy up, cost down)

Per-candidate measured cost/task and accuracy so far. The archive contains every nondominated vector; it is not ranked into one champion.

{objective_history}

### Full-validation interpretation

{full_validation_context}

When C{candidate_idx} has a measured full-validation row:

1. Compare its cost and accuracy with the seed, immediate parent, current, and other Pareto archive members.
2. Check whether the same-minibatch cost direction held at full-validation scale.
3. Attribute the aggregate movement only to mechanisms supported by the patch and available evidence; do not invent per-scenario validation behavior.
4. Record the conclusion through `--generalization-note` on the relevant scenario reflections. If C{candidate_idx} has no measured row, say that generalization is unmeasured rather than inferring it from prior candidates.

[...]

Never rewrite these machine-computed decisions from trace interpretation.

Separately classify whether the intended mechanism was exercised and worked:

- `supported`: the changed path ran, replaced the diagnosed waste, and the measured movement supports the mechanism.
- `partially_supported`: it worked on only part of the path or one objective.
- `unsupported`: the path did not run or did not replace the target behavior.
- `contradicted`: the path ran but added work or caused the opposite behavior.
- `unmeasured`: the current scenarios did not exercise the mechanism.

Also record causal attribution as `patch_caused`, `stochastic`, `mixed`, or `uncertain`. Mechanism verdict and causal attribution are analysis outputs; objective movement, train screening, and MOO status are machine inputs.

[...]

- **`quality_gain_cost_loss`**: Did extra work causally buy the quality gain? If development-evaluated, did it produce a nondominated admissible point?
- **`cost_gain_quality_loss`**: Did savings come from removing waste or from skipping required work? Did the development admit it?

[...]

There are four independent classifications on every command:

- `--status` is the exact-pass transition from the per-scenario table.
- `--movement-label` is the machine-computed scenario movement.
- `--mechanism-verdict` records whether the modified mechanism was exercised.
- `--causal-attribution` records whether behavior was patch-caused, stochastic,
  mixed, or uncertain.
\end{promptbox}
\section{Limitations and Future Work}
\paragraph{Evaluation cost and uncertainty.}
Repeated executions improve optimization evidence but increase
evaluation cost. Our Bayesian bootstrap captures variation among
executions on fixed screening tasks, rather than
uncertainty from task sampling. Future work could jointly adapt task selection and repetition counts to allocate evaluation effort
where promotion decisions remain uncertain.

\paragraph{Beyond Pareto-aware promotion.}
Our promotion rule provides one operational definition of a promising
intermediate candidate. Other local gates could preserve candidates
that improve particular task instances or exhibit novel behaviors,
even without favorable aggregate quality--cost changes.
An LLM-based gate could also use execution traces and optimization
history to judge whether a candidate warrants broader development
evaluation. Evaluating these alternatives
would clarify how to balance immediate gains, potential for later
improvement, and screening cost.

\end{document}